\documentclass{article} % For LaTeX2e

\usepackage[table]{xcolor}

\usepackage[final]{colm2025_conference}

\usepackage{microtype}
\usepackage{hyperref}
\usepackage{url}
\usepackage{booktabs}

\usepackage[utf8]{inputenc}
\usepackage{graphicx}

\usepackage{booktabs}
\usepackage{multirow}
\usepackage{tabulary}

\usepackage{subcaption}
\usepackage{adjustbox}
\usepackage{hyperref}

\usepackage{longtable}

\usepackage{times}
\usepackage{tipa}

\usepackage{rotating}
\usepackage{longtable}

\usepackage{lineno}

\usepackage{wrapfig}

\definecolor{darkblue}{rgb}{0, 0, 0.5}
\hypersetup{colorlinks=true, citecolor=darkblue, linkcolor=darkblue, urlcolor=darkblue}

\newcommand{\highlight}[2]{\colorbox[HTML]{#1}{#2}}

\definecolor{plotlyred}{RGB}{214, 39, 40}
\definecolor{plotlygreen}{RGB}{44, 160, 44}

\DeclareUnicodeCharacter{03B5}{\textepsilon}

\title{\textsc{unveiLing}: What Makes Linguistics Olympiad Puzzles Tricky for LLMs?}

\renewcommand\footnotemark{}

\author{Mukund Choudhary\textsuperscript{1,*,†}\thanks{\textsuperscript{*}Equal contribution. \textsuperscript{†}Correspondence to \texttt{mukund.choudhary@mbzuai.ac.ae}.}, KV Aditya Srivatsa\textsuperscript{1,*}, Gaurja Aeron\textsuperscript{2}, \\
\textbf{Antara Raaghavi Bhattacharya\textsuperscript{3,\#}\thanks{\textsuperscript{\#}Work done during an internship at MBZUAI.}, Dang Khoa Dang Dinh\textsuperscript{4,\#}, Ikhlasul Akmal Hanif\textsuperscript{5,\#}}, \\
\textbf{Daria Kotova\textsuperscript{1}, Ekaterina Kochmar\textsuperscript{1}, Monojit Choudhury\textsuperscript{1}} \\
\textsuperscript{1}Mohamed Bin Zayed University of Artificial Intelligence\ \textsuperscript{2}IIT Gandhinagar \\ \textsuperscript{3}Harvard University \ \textsuperscript{4}VinUniversity \ \textsuperscript{5}Universitas Indonesia
}

\begin{document}

\ifcolmsubmission
\linenumbers
\fi

\maketitle

\begin{abstract}

Large language models (LLMs) have demonstrated potential in reasoning tasks, but their performance on linguistics puzzles remains consistently poor. These puzzles, often derived from Linguistics Olympiad (LO) contests, provide a minimal contamination environment to assess LLMs' linguistic reasoning abilities across low-resource languages. In this work, we analyze LLMs' performance on 629 problems across 41 low-resource languages by labelling each with linguistically informed features to unveil weaknesses. Our analyses show that LLMs struggle with puzzles involving higher morphological complexity and perform better on puzzles involving linguistic features that are also found in English. We also show that splitting words into morphemes as a pre-processing step improves solvability, indicating a need for more informed and language-specific tokenisers. These findings thus offer insights into some challenges in linguistic reasoning and modelling of low-resource languages.

\end{abstract}

\section{Introduction}

% \begin{figure}
\begin{wrapfigure}{l}{0.45\textwidth}
    \centering
    \vspace{-0.4cm}
    \includegraphics[width=0.43\textwidth]{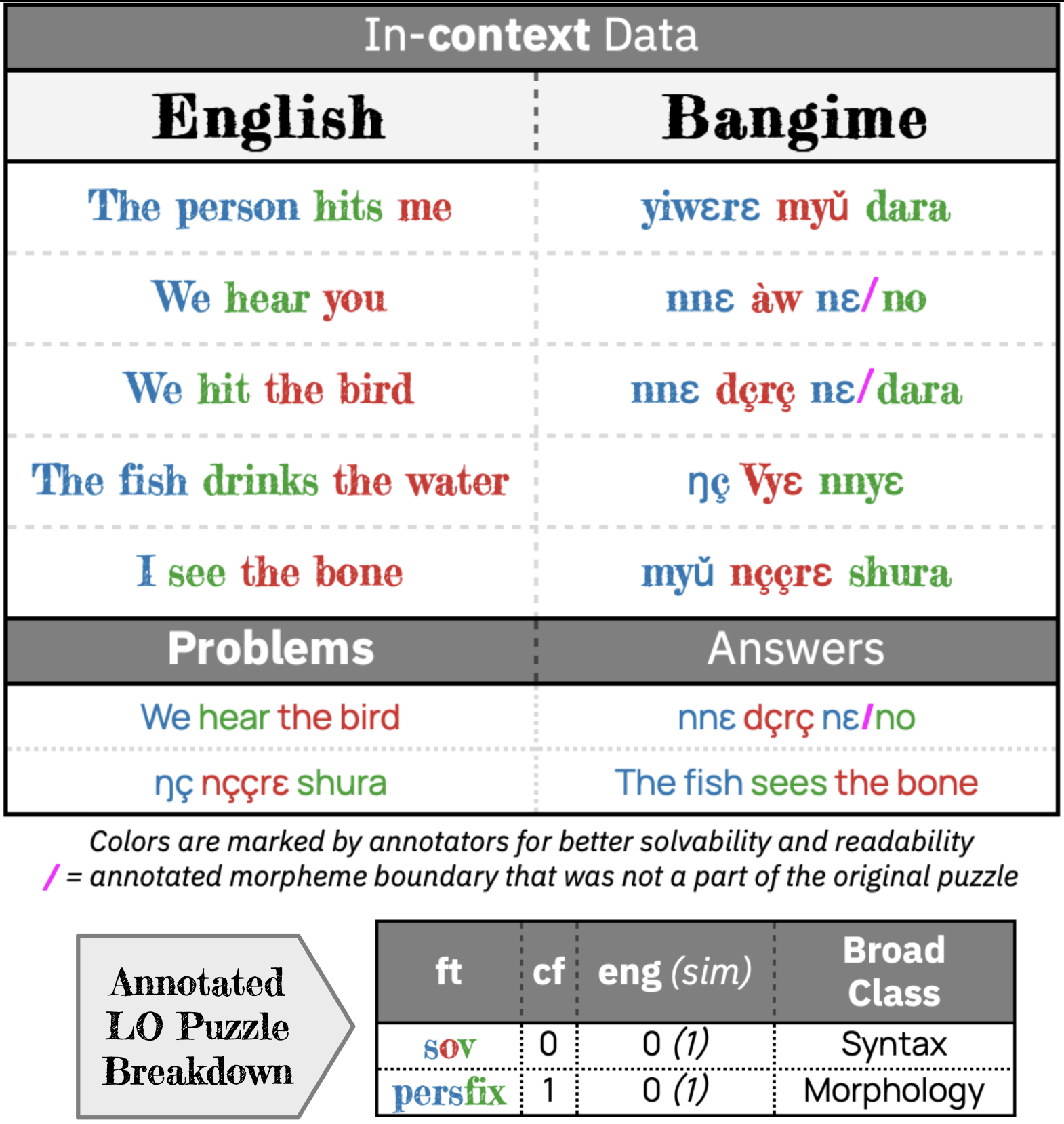}
    \caption{An LO puzzle with \textsc{unveiLing} annotations. In-context data are translation pairs, and problems are to be translated on the basis of the data above.}
    \vspace{-0.5cm}
    \label{fig:corr-plot}
\end{wrapfigure}
% \end{figure}

% can be removed
% LLMs show promise in various tasks involving reasoning in competitive coding \citep{shi2024can} or understanding low-resource languages (LRL) using linguistic descriptions \citep{tanzer2023benchmark}. However, an efficient way ahead to improve LLMs is through probing for weaknesses in linguistic reasoning \citep{waldis2024holmes}. Especially studying these abilities in LRL data where it is tougher to mimic training data (low leaks in the data \citep{huang2022large, bean2024lingoly}) requires crafting specific tests, which are difficult in English \citep{warstadt-etal-2020-blimp-benchmark} and trickier in others. Recent work tackles this with multilingual benchmarks derived from Linguistics Olympiads (LO) \citep{chi-etal-2024-modeling}. 

Human proficiency in analysing languages and linguistic reasoning contributes significantly to efficient learning from limited linguistic input. Following frameworks and goals established by \citet{mahowald2024dissociating, mcclelland2019extending, marcus2020next, beguvs2023large}, improvements in LLM metalinguistic processing capabilities can enhance general performance across diverse tasks. Linguistics Olympiad (LO) puzzles provide us with rich data to probe into these LLM abilities, especially in extremely low resource languages (LRL), as past research has shown \citep{chi-etal-2024-modeling}. Typically, LO puzzles have parallel data between an unknown language and corresponding forms in a language known to the solver \citep{bozhanov2013rosetta}. The solver (typically pre-collegiate level) is expected to deduce the mapping between the linguistic structures of the unknown and the known languages, which involves observing syntactic patterns, semantic correspondences, phonological rules, or a combination of these. We shall refer to them as features akin to linguistic features \citep{wals}. Fig. \ref{fig:corr-plot} shows an example of a puzzle set in Bangime (an unknown language) with the corresponding English (a known language) translations. The puzzles are similar to in-context few-shot learning tasks \citep{brown2020} where \textit{context} contains all the data needed to solve the \textit{problems}.

Generally, these are built from data in endangered languages or LRLs to ensure that participants at the Olympiads do not already know the patterns in the data. Therefore, we expect these languages to be absent from the LLM training data as well. This makes LO puzzles a challenging and minimal contamination setting to assess an LLM's linguistic reasoning abilities, and, for that reason, they have been incorporated into various benchmarks \citep{bbeh}.

% LO puzzles are self-sufficient Rosetta-stone style translation tasks \citep{bozhanov2013rosetta} between an LRL and English, accessible for pre-collegiate students. The solver is expected to deduce selected linguistic features from the patterns in a small set of paired translations (\textit{context}) and then applies them productively (like a native speaker) on incomplete pairs (\textit{problems}) (Fig.\ref{fig:corr-plot}). The puzzles are thus similar to in-context few-shot learning tasks \citep{brown2020} where \textit{context} contains all the data needed to translate the \textit{problems}. This makes LO puzzles a challenging and minimal contamination setting to assess an LLM's linguistic reasoning abilities and has been incorporated into benchmarks \citep{bbeh}. 

\begin{table*}[t!]
\centering
\resizebox{\textwidth}{!}{%
\begin{tabular}{lccccc}
\hline
\textbf{Dataset} & \textbf{Source} & \textbf{Puzzles (Problems)} & \textbf{Classes} & \textbf{Difficulty} & \textbf{Languages} \\ \hline
\begin{tabular}[c]{@{}l@{}}\textsc{modeLing}\\ \citep{chi-etal-2024-modeling}\end{tabular} & Author-made & 48 (272) & \begin{tabular}[c]{@{}c@{}}Noun-Adjective (40\%), Semantics (10\%), \\ Word Order (40\%), Possession (10\%)\end{tabular} & \begin{tabular}[c]{@{}c@{}}1 (3\%), 2 (34\%), 3 (23\%), \\ 4 (11\%), 5 (29\%)\end{tabular} & 19 \\ \hline
\begin{tabular}[c]{@{}l@{}}\textsc{PuzzLing}\\ \citep{csahin2020puzzling}\end{tabular} & \begin{tabular}[c]{@{}c@{}}National \\ Linguistics \\ Olympiads\end{tabular} & 96 (752) & -- & -- & 81 \\ \hline
\begin{tabular}[c]{@{}l@{}}\textsc{LingOly}\\ \citep{bean2024lingoly}\end{tabular} & \begin{tabular}[c]{@{}c@{}}UK \\ Linguistics \\ Olympiad\end{tabular} & 90 (1133) & \begin{tabular}[c]{@{}c@{}}Compounding, Morphology, \\ Numbers, Phonology, \\ Semantics, Syntax\end{tabular} & \begin{tabular}[c]{@{}c@{}}Breakthrough (6.5\%), Foundation (13.3\%), \\ Intermediate (13.5\%), Advanced (32.9\%), \\ Round 2 (33.7\%)\end{tabular} & 94 \\ \hline
\begin{tabular}[c]{@{}l@{}}\textsc{Linguini}\\ \citep{sanchez2024linguini}\end{tabular} & \begin{tabular}[c]{@{}c@{}}International \\ Linguistics \\ Olympiad\end{tabular} & 160 (894) & \begin{tabular}[c]{@{}c@{}}Sequence Transduction, \\ Fill-In Blanks, \\ Number Transliteration\end{tabular} & -- & 75 \\ \hline
\end{tabular}%
}
\caption{Summary of datasets with Linguistic Olympiad style puzzles. \% indicates proportion in data}
\label{tab:datasets-summary}
\end{table*}

Several researchers have used LO puzzles to study language models' reasoning abilities (summarized in Tab. \ref{tab:datasets-summary}). LLMs (while being impressive in application-based language tasks) perform consistently poorly in these Olympiads, whereas high school students can solve most of the puzzles fully.  While these studies show that LLMs struggle, they do not explain why. Since these puzzles reflect the complex interplay of linguistic features in natural languages, pinpointing the causes of poor performance is challenging. On the other hand, a feature-level analysis of LLM weaknesses can enhance our understanding of their linguistic reasoning and can provide insights into bridging the gap between LLM abilities and a core component of human cognitive flexibility to identify patterns, extract rules, and generalise from minimal data across unfamiliar languages.

% Research shows that LO puzzles are difficult for LLMs \citep{sanchez2024linguini} and machine translation systems \citep{csahin2020puzzling} by comparing them against human scores \citep{bean2024lingoly}, or expert-tagged difficulty \citep{chi-etal-2024-modeling}. However, research has not explored why models do not perform well or dissect the challenges at a puzzle level. As these puzzles are an involved interplay of linguistic features of the language they are based on, it is difficult to study them by breaking them down into features. We address this by annotating puzzles with \textit{World Atlas of Language Structures}' (WALS) \citep{wals} features and finding correlations with LLM scores.

To answer the research question of \textit{what linguistic features make LO puzzles challenging for LLMs}, we have annotated 64 puzzles (629 problems, 41 LRLs) from existing datasets (Sec. \ref{sec:data-refinement}) with 50 linguistic features (from WALS,\footnote{World Atlas of Languages by \cite{wals}} also including other attributes such as similarity to English -- see Sec. \ref{sec:feat}). We discover that LLM performance (Sec. \ref{sec:exp}) is significantly negatively correlated with the number of morphological features of a language present in an LO puzzle (Sec. \ref{sec:morphanal}) and the number of features in a puzzle that have a low coverage of exemplifying data in the puzzle (Sec. \ref{sec:cfanal}). LLM scores also positively correlated with the average number of features in a puzzle that are present in English (Sec. \ref{sec:enganal}). Further experiments showed that breaking down words in LRL into morphemes in morphologically complex puzzles helped LLMs (Sec. \ref{sec:morphpoc}), seemingly indicating the need for more language-specific handling of tokenisation of words in LRL. Findings from the study could help inform future LLM-based research on sample-efficient LRL learning, metalinguistic reasoning, and more.

\section{Setting the Stage for \textsc{unveiLing}}
\label{sec:datafeat}
In this section, we set the stage for unveiling LLM weaknesses in solving LO puzzles by:
\begin{enumerate}
    \item Curating data and finding unseen puzzles.
    \item Establishing an annotation scheme to break down all puzzles using linguistic features.
    \item Obtaining LLM responses to evaluate performance against the feature annotations.
\end{enumerate}

\subsection{Data}
\label{sec:data-refinement}

\paragraph{Existing Datasets} An overview of the four datasets that are based on LO-style puzzles can be found in Tab. \ref{tab:datasets-summary}. All of these reveal that LLMs are poor at solving LO problems based on Exact-Match Accuracy, which measures the percentage of problems per puzzle where the predicted final answer exactly matches the ground truth. LLMs performed the best (40-60\%) in solving puzzles from {\sc modeLing}~\citep{chi-etal-2024-modeling} (author-made, easier LO-style puzzles) and Foundation-level (younger students) puzzles in {\sc LingOly}~\citep{bean2024lingoly}, while performing the worst (25\%) on International Linguistics Olympiad (IOL) puzzles from {\sc Linguini}~\citep{sanchez2024linguini}. 

\paragraph{Filtering out Solvable Puzzles} LLMs struggle with {\sc Linguini}'s set of IOL puzzles as they are tougher and are designed for participants who have more experience with LO puzzles. Thus we only consider puzzles from {\sc modeLing} (M) and {\sc LingOly} (L). To maintain uniformity across datasets and select the simplest tasks, we retained only Rosetta-Stone style \citep{bozhanov2013rosetta} puzzles which had paired translations in context and problems that required participants to either translate from an LRL to English or vice versa. This filtering step led to an initial pool of puzzles consisting of all 48 M puzzles\footnote{All M puzzles were first solved and checked by the authors for inconsistencies (L puzzles were already checked by UKLO participants). Issues included insufficient data, absence of affixes in answers, etc. A detailed account of minor changes made to the data is provided in App. \ref{app:modeling-corrections}.} and 45 L puzzles out of 90.

\begin{figure*}
    \centering
    \includegraphics[width=\textwidth]{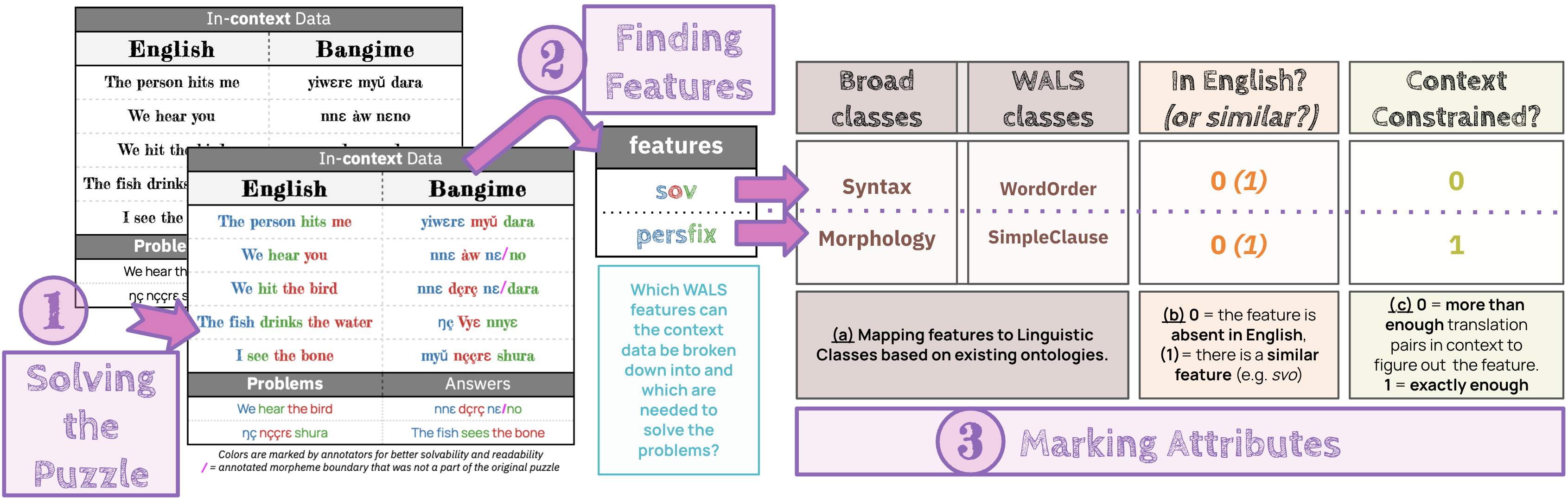}
    \vspace{-0.4cm}
    \caption{\textsc{unveiLing} annotation pipeline and attributes marked in a sample puzzle from the dataset.}
    \vspace{-0.5cm}
    \label{fig:annotation_pipeline}
\end{figure*}

\paragraph{Contamination Checks and Pruning} 

{\sc ModeLing} (M) and {\sc LingOly} (L) employ measures to account for data contamination, i.e., being privy to the exact puzzles or being familiar with the languages involved, which may lead to inflated results. M puzzles are based on extremely low-resource languages that were reported to have 0\% data leakage to OpenAI models up to GPT-4. \citet{bean2024lingoly} (L) report all analyses with a \texttt{NoContext} baseline for querying problems without the puzzle in-\textit{context} data and improvements over it. 

\begin{wrapfigure}[12]{t}{0.5\textwidth}
    \centering

    \vspace{-2mm}
    
    \begin{subfigure}[b]{0.45\textwidth}
        \centering
        \includegraphics[width=\linewidth]{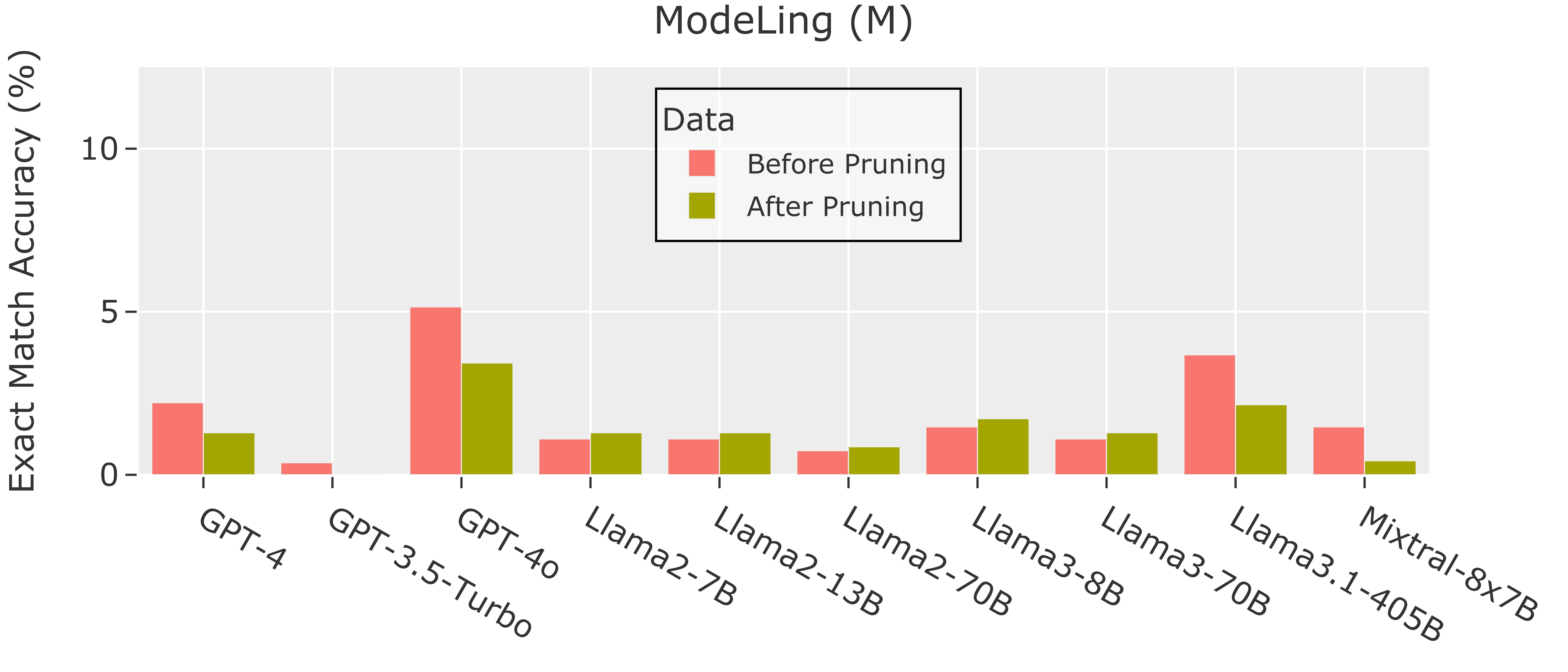}
        \label{fig:sub1}
    \end{subfigure}
    % \hfill
    \vspace{-10mm}
    
    \begin{subfigure}[b]{0.45\textwidth}
        \centering
        \includegraphics[width=\linewidth]{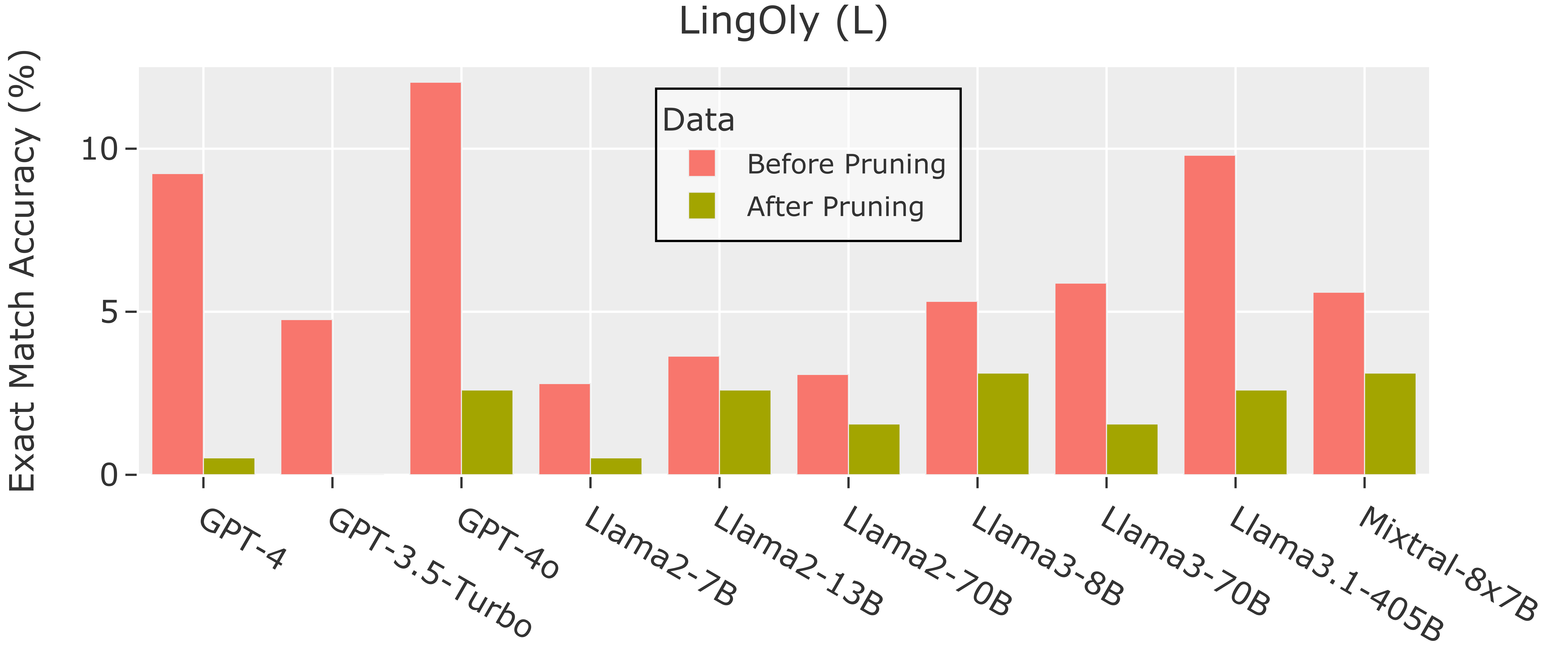}
        \label{fig:sub2}
    \end{subfigure}
    
    \vspace{-8mm}
    
    \caption{Results on the \texttt{NoContext} prompt to find puzzles with potential contamination.}
    \label{fig:no_context}
\end{wrapfigure}

We employ a similar measure by using a \textbf{\texttt{NoContext}} prompt (App. \ref{app:querying-details}) followed by a manual inspection of all model responses and drop the contaminated puzzles from both datasets (Fig. \ref{fig:no_context}): 7 out of 48 M puzzles (answered by the GPT-4o\footnote{This model was not public at the time of \textsc{modeLing}'s publication.} without any in-context data) and 22 out of 45 L puzzles were pruned. This is expected as there are more high-resourced languages (e.g., Romani and Nepali) in the L puzzles set, and corresponding puzzles are available online via UKLO website. 

The final dataset contains 64 puzzles (629 problems):\footnote{Each \textit{puzzle} is made of multiple translation \textit{problems}, which are evaluated independently.} 41 (272) M + 23 (357) L.\footnote{Remaining contamination ($\leq$ 4\%) was where models matched answers by picking random words from both languages, as the solution could be pieced together with a random sequence of strings from the questions.}

\subsection{Feature \& Attribute Annotations}

\label{sec:feat}

\paragraph{Background \& Motivation} LLM performance explanations in literature have been limited to using broad external difficulty metrics (e.g. participant scores or Likert ratings -- see Tab. \ref{tab:datasets-summary}). As LLMs might model language differently \citep{mcoy, warstadt-etal-2020-blimp-benchmark}, such measures conflate linguistic reasoning with other puzzle-specific quirks and offer low interpretability.

A more comprehensive probe of LLM linguistic abilities in English was done by \citet{warstadt-etal-2020-blimp-benchmark} and \citet{waldis2024holmes}. They craft linguistic textbook-style tasks to find that linguistic competence correlates with the model size, but also that different features have different learning curves and instruction tuning negatively influences performance and natural quality -- i.e., they mimic linguistic features instead of generalising productively like a native speaker. They also find that formal linguistic abstractions like {\em syntax} in English are more easily modelled by LLMs as compared to less explicitly lexically or structurally represented abstractions like {\em pragmatics}.

LO puzzles offer an opportunity to conduct similar studies on LLM abilities in LRLs. However, as they are based on real language (rather than artificially controlled) data, each puzzle involves simultaneously changing features from different linguistic abstractions like {\em phonology}, {\em semantics}, and so on (Fig. \ref{fig:corr-plot}).

Thus, the following sections describe how we analyse LO puzzles by building on top of \cite{bozhanov2013rosetta}'s insight that LO puzzles are crafted “to be solved by deducing linguistic patterns, with other sources of technical complexity … not making them linguistically interesting”. As Fig. \ref{fig:annotation_pipeline} indicates, we break them down into fine-grained, linguistically grounded proxies for complexity, which we call \textit{``Attributes''} from this point in the paper. Attributes include 50 linguistic \textit{``Features''} from abstractions like {\em morphology}, {\em syntax}, etc., and other derived attributes such as how many features are similar to English features, and how much evidence is redundant to derive features from.

\paragraph{Methodology} 2 ex-LO participants -- with training in Linguistics and experience in LO puzzle design -- independently annotated all puzzles with linguistic features from WALS and classified them in three ways following the bottom-up fashion \citep{cristani2005survey} as is outlined below. For an overview of the annotation pipeline, see Fig. \ref{fig:annotation_pipeline}; the annotation guidelines are presented in App. \ref{app:ann}).

\begin{wraptable}[19]{r}{0.5\textwidth}
\centering
\vspace{-1.25cm}
\resizebox{\linewidth}{!}{%
\begin{tabular}{@{}llcc@{}}
\toprule
\textbf{Attribute} & \textbf{Classes} & \textbf{$\mu$} & \textbf{$\sigma$} \\ \midrule
\multirow{4}{*}{\begin{tabular}[c]{@{}l@{}}Proportion of \\ Broad Attributes \\ per Puzzle\end{tabular}} & \texttt{b\_Syntax} & 0.519 & 0.381 \\ \cmidrule(l){2-4} 
 & \texttt{b\_Morphology} & 0.307 & 0.322 \\ \cmidrule(l){2-4} 
 & \texttt{b\_Semantics} & 0.132 & 0.243 \\ \cmidrule(l){2-4} 
 & \texttt{b\_Phonology} & 0.042 & 0.11 \\ \midrule
\multirow{9}{*}{\begin{tabular}[c]{@{}l@{}}Proportion of \\ WALS Attributes \\ per Puzzle\end{tabular}} & \texttt{w\_WordOrder} & 0.437 & 0.399 \\ \cmidrule(l){2-4} 
 & \texttt{w\_ComplexSentences} & 0.004 & 0.023 \\ \cmidrule(l){2-4} 
 & \texttt{w\_NominalSyntax} & 0.096 & 0.186 \\ \cmidrule(l){2-4} 
 & \texttt{w\_SimpleClauses} & 0.131 & 0.192 \\ \cmidrule(l){2-4} 
 & \texttt{w\_VerbalCategories} & 0.03 & 0.078 \\ \cmidrule(l){2-4} 
 & \texttt{w\_NominalCategories} & 0.164 & 0.244 \\ \cmidrule(l){2-4} 
 & \texttt{w\_Morphology} & 0.075 & 0.181 \\ \cmidrule(l){2-4} 
 & \texttt{w\_Lexicon} & 0.023 & 0.131 \\ \cmidrule(l){2-4} 
 & \texttt{w\_Phonology} & 0.039 & 0.109 \\ \midrule
\multicolumn{2}{l}{\begin{tabular}[c]{@{}l@{}}Proportion of Data Constrained \\ Features per Puzzle (\texttt{cf})\end{tabular}} & 0.474 & 0.435 \\ \midrule
\multicolumn{2}{l}{\begin{tabular}[c]{@{}l@{}}Proportion of Features \\ also Present in English per Puzzle (\texttt{eng})\end{tabular}} & 0.298 & 0.36 \\ \midrule
\multicolumn{2}{l}{\begin{tabular}[c]{@{}l@{}}Proportion of Features \\ Similar to English per Puzzle (\texttt{eng sim})\end{tabular}} & 0.498 & 0.381 \\ \bottomrule
\end{tabular}%
}
\vspace{-0.25cm}
\caption{Summary Statistics of all attributes -- means ($\mu$) and standard deviations ($\sigma$).}
\label{tab:attr_stats}
\end{wraptable}

\subsubsection{Solving Puzzles}
All puzzles were first solved independently by the annotators. They also marked the context data used to translate the problems to help with the next steps.

\subsubsection{Finding Features (\texttt{ft})}
Broken-down contexts were then used to extract features. A feature here (defined by \citet{wals}, similar to \citet{warstadt-etal-2020-blimp-benchmark}'s \textit{phenomenon}) is a structural property of language that describes one aspect of cross-linguistic diversity. Annotations were based on a set of 50 WALS features\footnote{List of features can be found at App. \ref{sec:app-feat_list}.} comprising various grammatical and semantic affixation, vowel harmony, abstract concepts like moiety, and more. Feature description with examples and citations can be found in App. \ref{app:featdefs}.\footnote{Code and annotations are publicly available at \url{https://github.com/mukundc2k/unveiling}}

\subsubsection{Marking for Attributes}
\paragraph{(a) Linguistic Attributes}
We cluster features into linguistic abstractions defined by WALS and UKLO\footnote{\href{https://www.eva.mpg.de/lingua/research/featurelist.php}{A quick reference for WALS classes} and \href{https://www.uklo.org/technical-information/}{for UKLO classes}} for a more generalized but still linguistically grounded analysis of LLM abilities:
\begin{itemize}
    \item \textbf{WALS classes (\texttt{w\_})}:
    \begin{itemize}
        \item \texttt{Nominal/VerbalCategories}: grammatical features relating to nouns and verbs such as case, tense, and aspect.
        \item \texttt{WordOrder}: Order of frequent constituents in a noun phrase (like adjectives) or a sentence (like Subject-Object).
        \item \texttt{Phonology}: Features based on units of sound, tones, vowel harmony, and so on.
        \item \texttt{Lexicon}: Lexical items representing number systems, alienability, and so on.
        \item \texttt{ComplexSentences}: Relative clauses and other multi-clause constructions.
        \item \texttt{NominalSyntax/SimpleClauses}: Order of minor constituents of a noun phrase (such as possessive markers) or a clause (such as voice marker).
        \item \texttt{Morphology}: Reduplication and other features that use morphemes, apart from nominal/verbal affixes above.
    \end{itemize} 
    \item \textbf{Broad classes (\texttt{b\_})}: \texttt{Syntax} (phrasal/clausal rules), \texttt{Phonology} (sound units), \texttt{Morphology} (subword units), \texttt{Semantics} (meaning units).
\end{itemize}

\paragraph{(b) Similarity to English (\texttt{eng}):} To unveil biases from typological distance from English, we marked how many features in a puzzle were present in English or exemplified similar behavior (\texttt{sim}).

\paragraph{(c) Constrained features (\texttt{cf}):} It is generally easier for a solver to uncover a feature and its rules in an LO puzzle when it appears in different contexts with redundant minimal pairs of translations. A puzzle is tougher if another feature changes in the same evidence pairs. Thus to unveil if LLMs face similar difficulties, we mark how many features have exactly enough translation pairs to uncover the respective feature from the context data.\footnote{There can never be fewer than enough contexts by design.} 

\begin{wrapfigure}[]{t}{0.5\textwidth}
    \centering
    \vspace{-0.6cm}
    \includegraphics[width=\linewidth]{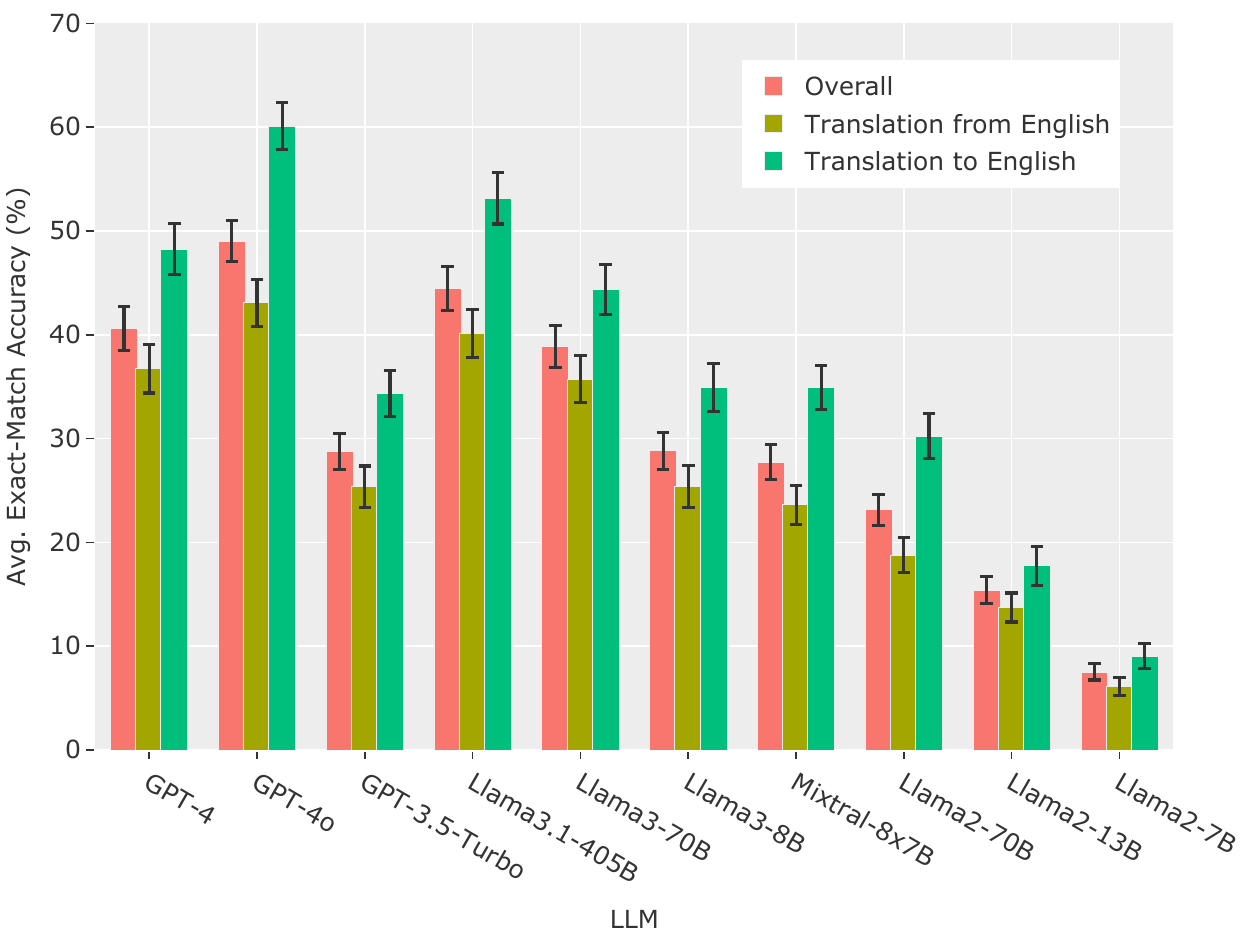}
    \vspace{-0.6cm}
    \caption{Benchmarking results -- overall and language direction-specific average Exact-Match accuracies for all models across prompt settings.}
    \vspace{-1cm}
    \label{fig:benchmain}
\end{wrapfigure}

\subsubsection{Resolving Annotation Differences}
Finally, after comparing the independent annotations, we found that the annotators agreed almost perfectly on which WALS features were present in a puzzle (Cohen's $\kappa$=0.893), substantially on how many constrained features were present per puzzle ($\kappa$=0.631), and on the count of features per puzzle that were present in English as well ($\kappa$=0.761).

They resolved the disagreements by relabeling the puzzles with newly agreed-upon reasoning. Specifically, for finding constrained features, they independently came up with pairs in contexts relevant to uncovering the feature and agreed on a union. For English similarity, they referred to WALS and other research in English linguistics (cited for all features in App. \ref{app:featdefs}).

Tab. \ref{tab:attr_stats} provides a statistical summary of all attributes used for analyses (Sec. \ref{sec:featanl}).

\subsection{Benchmarking Experiments}
\label{sec:exp}

We evaluate 10 LLMs:\footnote{Models were selected to be representative of various parameter sizes, families, and to contain both open \& closed models for balance and comparability. Refer to App. \ref{app:querying-details} for more details.} {\tt GPT} {\tt4}, {\tt4o}, {\tt3.5}{\tt-Turbo} \citep{4o}, {\tt Llama2} {\tt7B}, {\tt13B}, {\tt70B} \citep{llama2}, {\tt Llama3} {\tt8B}, {\tt70B}, {\tt3.1-405B} \citep{llama3}, and {\tt Mixtral 7x8B} \citep{mixtral}. These were prompted with 6 styles and strategies from previous work:\footnote{\textsc{m} marked prompts are borrowed from \citet{chi-etal-2024-modeling}, \textsc{l} is borrowed from \citet{bean2024lingoly}.}

\begin{itemize}
    \item \textbf{Null}: prompt to gauge model performance without explicit instructions.
    \item \textbf{Minimal}\textsuperscript{\textsc{m}}: similar to Null but with task description.
    \item \textbf{Hand-Tuned}\textsuperscript{\textsc{m}}: prompt fine-tuned by an IOL medalist.
    \item \textbf{Basic CoT}\textsuperscript{\textsc{m}}: Chain-of-Thought prompt asking the model to answer step-by-step.
    \item \textbf{Full CoT}\textsuperscript{\textsc{m}}: similar to the above but with an example of reasoning in a dummy puzzle.
    \item \textbf{LingOly}\textsuperscript{\textsc{l}}: A more detailed prompt similar to the Minimal prompt above.
\end{itemize}

We find that our results on the curated dataset are in line with existing literature (Fig. \ref{fig:benchmain}):

\begin{itemize}
    \item Larger and closed-source models are better \citep{srivastava2023beyond}. Smaller models ({\tt Mixtral-8x7B} and {\tt Llama3-8B}) outperform larger ones ({\tt GPT3.5-Turbo}) rarely, which can be explained by \citet{waldis2024holmes}'s finding that linguistic skills vary depending on architecture and tuning parameters.
    \item Models have a translation direction bias and score better when translating from the low-resource language {\em to} English compared to translating {\em from} English \citep{csahin2020puzzling}.
\end{itemize}

For a more detailed breakdown and other observations, please refer to App. \ref{app:bench}. Note that we use a model's average performance across prompts for the following analyses.

\begin{wrapfigure}[18]{t}{0.45\textwidth}
    \centering
    \includegraphics[width=\linewidth]{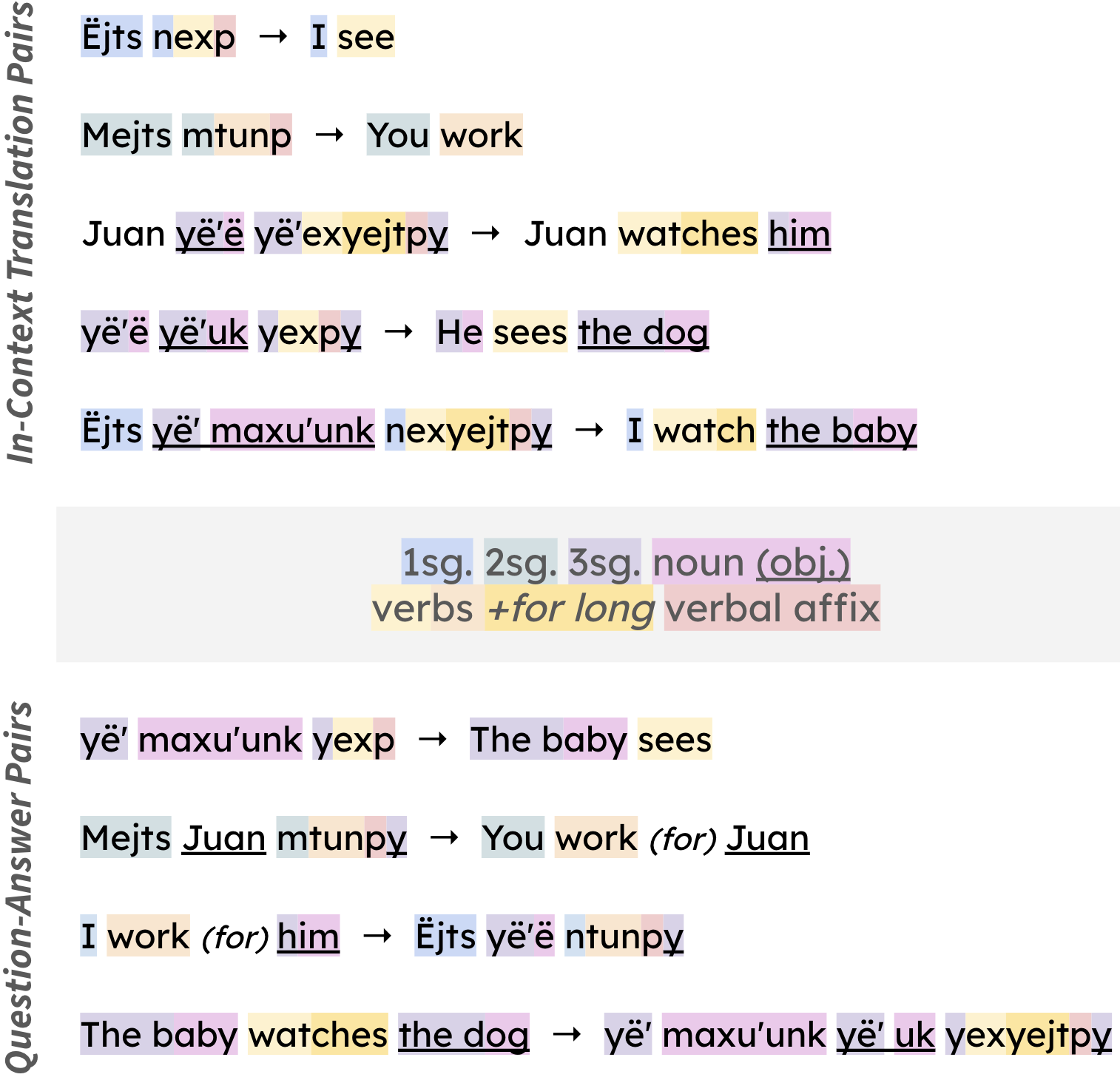}
    \caption{An annotated Ayutla puzzle. Colors mark equivalent units.}
    \label{fig:subfig2}
\end{wrapfigure}

\section{Analyses \textsc{unveiL} that...}
\label{sec:featanl}

To unveil patterns behind LLM performances we compute the correlations between the attributes and LLM exact match scores across prompts. We base our analyses only on significant correlations.

\subsection{Morphologically complex puzzles are difficult for LLMs}
\label{sec:morphanal}

We find that the count of morphological features (\texttt{b\_Morphology}) in an LO puzzle strongly predicts how well a model can solve the puzzle. Tab. \ref{tab:feat_vs_corr_main_paper}\footnote{"*", "**", and "***"  imply that the corresponding p-value is less than 0.05, 0.01, and 0.001, respectively.} shows that higher morphological feature count correlates with lower exact match scores across all LLMs (all significant p-values $\leq .001$). These scores are averaged over all prompting strategies; however, this observation holds over most individual model-prompt combinations as well. For all combinations, please refer to Tab. \ref{tab:feat_vs_corr_full} in Appendix.

\begin{wrapfigure}[20]{}{0.37\textwidth}
    \centering
    \vspace{-2mm}
    \includegraphics[width=\linewidth]{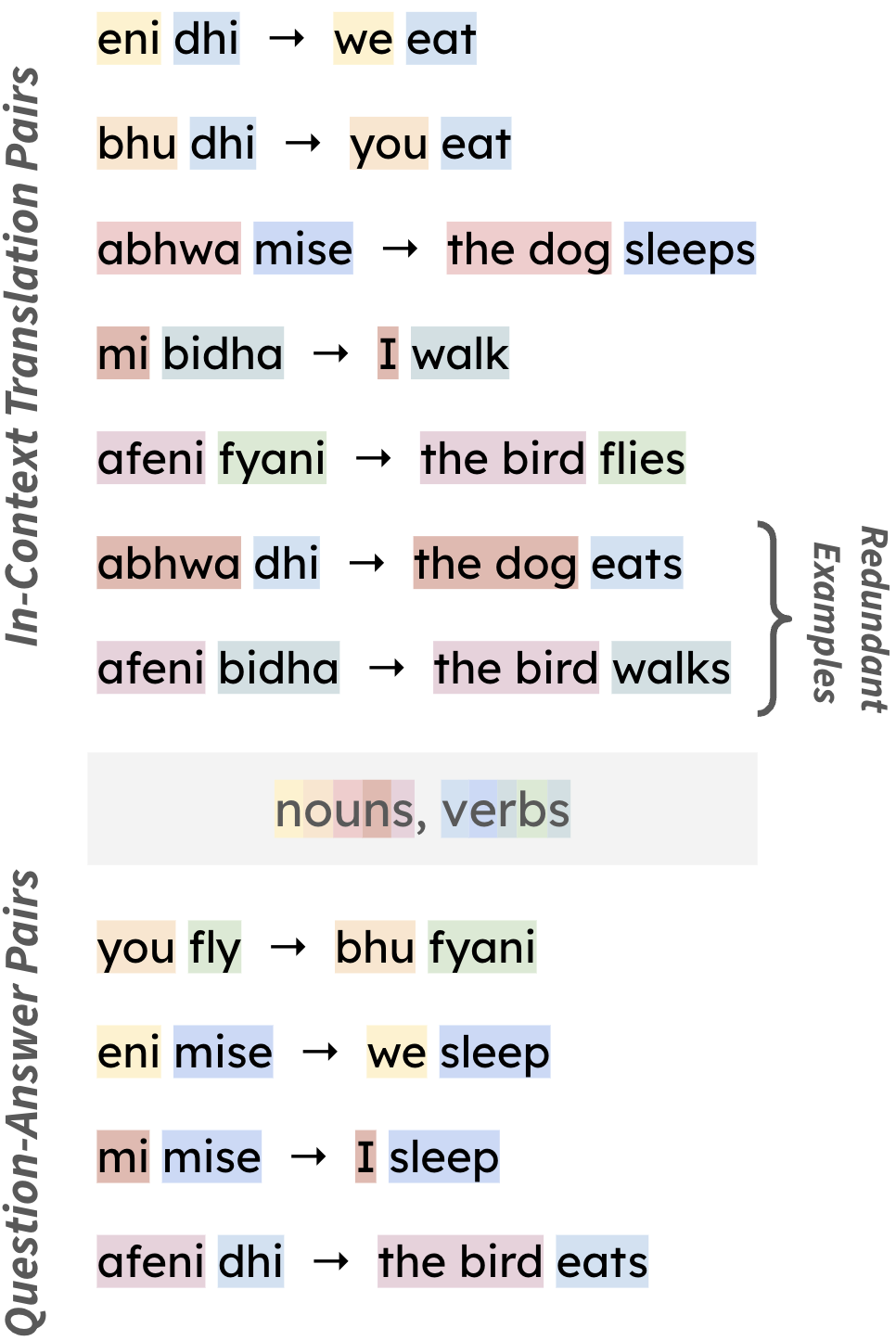}
    \caption{Annotated Engenni Puzzle}
    % \vspace{-10mm}
    \label{fig:subfig1}
\end{wrapfigure}

To provide some illustrative examples, a puzzle in Ayutla Mixe (Mixe-Zoque family, Fig. \ref{fig:subfig2}) encoded many morphological features, and no LLMs solved more than one question. In contrast, Bangime (language isolate from Mali, Fig. \ref{fig:corr-plot}) had only one such feature (affix for person), most LLMs solved all the problems except for the one that involved the affix. Finally, a puzzle with no morphological complexity, based on Engenni (Niger-Congo family, Fig. \ref{fig:subfig1}), was solved fully by all models.

Note that smaller and older models that, on average, score lower across all puzzles also showed lower correlations with \texttt{b\_Morphology}. Accurate answers by these models were mostly random or chance matches and were not consistent across questions in the same puzzle (better scoring models would either consistently not apply a morphological rule or use it productively across answers).

In Sec. \ref{sec:morphpoc}, we actively test our hypothesis of LLMs being weak in splitting morphemes into the right tokens in morphologically richer languages. Results show that simply making morpheme boundaries explicit increases the scores in better-performing models. 
% Finally, \texttt{w\_NominalCategories} and \texttt{w\_SimpleClauses} have a huge overlap with \texttt{b\_Morphology} and show similar trends.

\begin{table*}
\centering
\vspace{-0.55cm}
\resizebox{\textwidth}{!}{%
\begin{tabular}{@{}lcccccccccc@{}}
\toprule
\multicolumn{1}{c}{\textbf{}} & \textbf{\begin{tabular}[c]{@{}c@{}}Llama2\\ (7B)\end{tabular}} & \textbf{\begin{tabular}[c]{@{}c@{}}Llama2\\ (13B)\end{tabular}} & \textbf{\begin{tabular}[c]{@{}c@{}}Llama2\\ (70B)\end{tabular}} & \textbf{\begin{tabular}[c]{@{}c@{}}Mixtral\\ (8x7B)\end{tabular}} & \textbf{\begin{tabular}[c]{@{}c@{}}Llama3\\ (8B)\end{tabular}} & \textbf{\begin{tabular}[c]{@{}c@{}}Llama3\\ (70B)\end{tabular}} & \textbf{\begin{tabular}[c]{@{}c@{}}Llama3.1\\ (405B)\end{tabular}} & \textbf{\begin{tabular}[c]{@{}c@{}}GPT\\ (3.5-Turbo)\end{tabular}} & \textbf{\begin{tabular}[c]{@{}c@{}}GPT\\ (4o)\end{tabular}} & \textbf{\begin{tabular}[c]{@{}c@{}}GPT\\ (4)\end{tabular}} \\ \midrule
\textbf{Average Exact-Match Score} & 0.074 & 0.143 & 0.224 & 0.271 & 0.269 & 0.381 & 0.405 & 0.248 & 0.488 & 0.408 \\ \midrule
Count of \texttt{b\_Morphology} & -0.219\textsuperscript{***} & -0.348\textsuperscript{***} & -0.496\textsuperscript{***} & -0.546\textsuperscript{***} & -0.522\textsuperscript{***} & -0.583\textsuperscript{***} & -0.572\textsuperscript{***} & -0.483\textsuperscript{***} & -0.637\textsuperscript{***} & -0.553\textsuperscript{***} \\ \midrule
Average of \texttt{eng} & 0.031 & 0.053 & 0.388\textsuperscript{**} & 0.3\textsuperscript{*} & 0.491\textsuperscript{***} & 0.364\textsuperscript{**} & 0.276\textsuperscript{*} & 0.399\textsuperscript{**} & 0.258\textsuperscript{*} & 0.381\textsuperscript{**} \\ \midrule
Count of \texttt{cf} & -0.144 & -0.198 & -0.448\textsuperscript{***} & -0.466\textsuperscript{***} & -0.467\textsuperscript{***} & -0.55\textsuperscript{***} & -0.592\textsuperscript{***} & -0.543\textsuperscript{***} & -0.572\textsuperscript{***} & -0.547\textsuperscript{***} \\ \bottomrule
\end{tabular}%
}
\vspace{-0.25cm}
\caption{Correlation values (Pearson) between exact-match scores and attribute value aggregates per puzzle for selected attributes against all models across prompts.}
\label{tab:feat_vs_corr_main_paper}
\end{table*}

\subsection{LLMs have an English bias}
\label{sec:enganal}

We find that a puzzle's average number of features that are present in English (\texttt{avg\_eng}) strongly predicts how well an LLM scores on that puzzle. Tab. \ref{tab:feat_vs_corr_main_paper} shows that a higher average similarity positively correlates with higher exact match scores. Note that the average model-prompt combinations have a slightly lower significance level (p-value $\leq .01$); however, many individual model-prompt combinations are consistent with this observation (Tab. \ref{tab:feat_vs_corr_full}).

\begin{wrapfigure}[11]{}{0.3\textwidth}
    \centering
    \vspace{-0.6cm}
    \includegraphics[width=\linewidth]{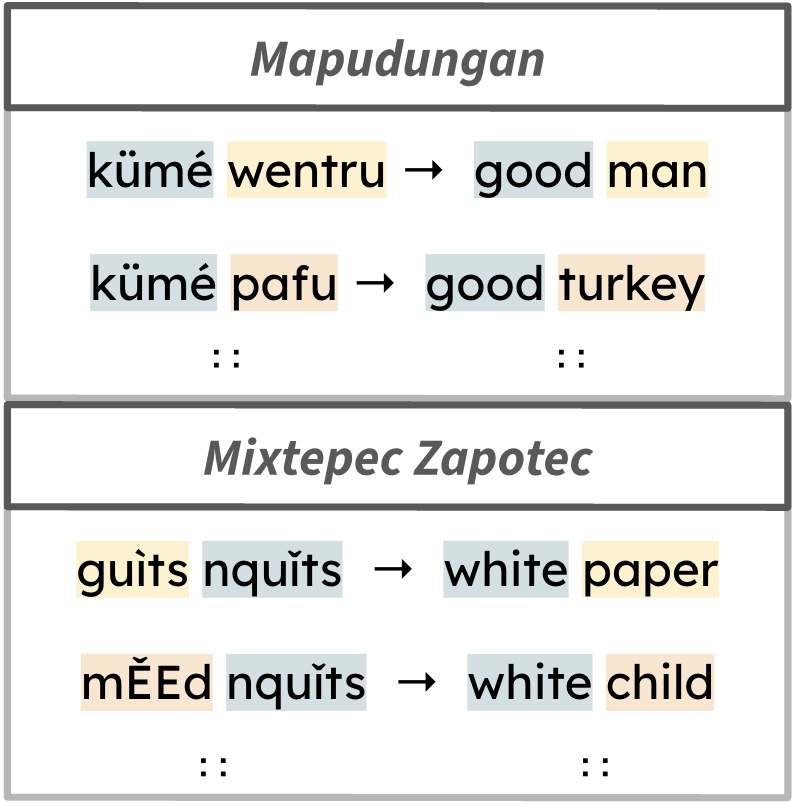}
    \vspace{-0.5cm}
    \caption{Sample of context examples of two puzzles.}
    \vspace{-0.5cm}
    \label{fig:eng-ex}
\end{wrapfigure}

To provide some illustrative examples (Fig. \ref{fig:eng-ex}), a puzzle in Mixtepec Zapotec (Oto-Manguean family) required the solver to figure out Noun-Adjective order (opposite to English); however, LLMs could not solve more than 30\% of the questions in this puzzle. In contrast, Mapudungan (Araucanian family) had the same setup as Mixtepec Zapotec, except that the word order in noun-adjective phrases in this language is the same as in English, so most LLMs fully solved this puzzle.

Note that most correlations are in the same range and consistent across prompts and models, showing that no specific model type is more or less biased towards English feature-like puzzles than others.

\subsection{More constrained evidence for a feature makes a puzzle difficult for LLMs}
\label{sec:cfanal}

We find that the count of features with exactly enough pairs needed to uncover them (\texttt{cf}) in an LO puzzle strongly predicts how well a model can solve the puzzle. Tab. \ref{tab:feat_vs_corr_main_paper} shows that higher exact feature count correlates with lower exact match scores across most LLMs (all significant p-value $\leq .001$ and negative correlations). Note that these scores are averaged over different prompting strategies however, this observation holds over most individual model-prompt combinations as well (Tab. \ref{tab:feat_vs_corr_full}).

The Engenni puzzle (Fig. \ref{fig:subfig1}) shows that the last two examples in the puzzle are redundant and are not required to uncover either the feature (Subject-Verb order) or vocabulary. Thus, there is more than enough evidence to figure out all features. LLMs scored higher on such puzzles -- i.e., puzzles where both annotators found ample evidence for features in them.

Note that smaller and older models lack these correlations due to low performance in general and due to not being able to answer any of the more data-constrained puzzles. Further, Null and Minimal prompts (which did not explain the task) weaken correlations for other mid-sized models.

\subsection{Other Observations}

Some other statistics and significant correlations that were only localised to certain model prompt combinations are summarised below.\footnote{Other attributes that were not mentioned in the above analyses were highly correlated with the ones discussed and/or were not represented in more than half of the dataset, to make any generalised claims.}

\texttt{w\_WordOrder} showed significant and negative correlations with the smaller and older models ({\tt Llama2-7B} \& {\tt 13B}) for specific prompts. Note that these are also the models that perform poorly overall and show no correlation with other attributes. {\em Semantics-} and {\em phonology}-based attributes across the two classification systems were scarce in the dataset, leading to low significance scores. However, at a slightly lower significance level (p-value $\leq .01$), the models from OpenAI showed a negative correlation, implying that puzzles with more {\em semantics-} or {\em phonology}-based features were tougher for OpenAI models (consistent with \citet{waldis2024holmes, warstadt-etal-2020-blimp-benchmark}). We also extended our experimental results to {\tt DeepSeek-R1}, a model tuned for Inference-Time Compute \citep{deepseekai2025deepseekr1incentivizingreasoningcapability} -- see Appendix \ref{app:subsec-r1-scores} for more details.

\begin{wrapfigure}{}{0.5\textwidth}
    \centering
    \vspace{-1.4cm}
    \includegraphics[width=\linewidth]{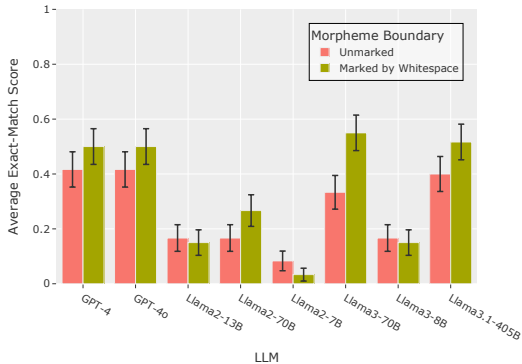}
    \vspace{-0.5cm}
    \caption{Performance change in models upon making morpheme boundaries explicit}
    \vspace{-0.5cm}
    \label{fig:poc-scores}
\end{wrapfigure}

\subsection{Breaking morphemes down helps LLMs}
\label{sec:morphpoc}

Insightful findings from Sec. \ref{sec:morphanal} unveiled that the morphological complexity of an LO puzzle strongly predicted LLM scores. We further probe a subset of 9 puzzles of varying morphological complexity and feature types on which models did not perform well.

\begin{wraptable}[15]{r}{0.5\textwidth}
\centering
\vspace{-0.3cm}
\resizebox{0.8\linewidth}{!}{%
\begin{tabular}{@{}lc@{}}
\toprule
\textbf{Puzzle} & \textbf{Exact-Match Score (+Gain)} \\ \midrule
\textit{L\_Ilokano} & 0.333 (+0.333) \\ \midrule
\textit{L\_Karelian} & 0.769 (+0.154) \\ \midrule
\textit{L\_Lardil} & 0.333 (+0.167) \\ \midrule
\textit{M\_Ayutla\_Mixe} & 0.0 (+0.0) \\ \midrule
\textit{M\_Bangime\_3} & 0.4 (+0.2) \\ \midrule
\textit{M\_Bangime\_5} & 0.6 (+0.2) \\ \midrule
\textit{M\_Guugu\_Yimithir} & 0.7 (+0.2) \\ \midrule
\textit{M\_Kutenai} & 1.0 (+0.6) \\ \midrule
\textit{M\_Totonac} & 0.333 (+0.167) \\ \bottomrule
\end{tabular}%
}
\caption{{\tt Llama3-70B} puzzle-wise score gain on marking morpheme boundaries with whitespaces}
\label{tab:poc_qwise_gains}
\end{wraptable}

As illustrated with pink bars in Fig. \ref{fig:corr-plot} and with boxes in Fig. \ref{fig:subfig2}, we first annotate the morpheme boundaries in the context (based on the language's grammar (App. \ref{app:featdefs}) and answers), problems, and answer data of these puzzles.

Then we run inference using selected models (with prompt styles that led to best scores respectively) again, with two variations of morpheme boundary representations: whitespace-separated (which made morpheme boundaries the same as the word boundaries), and @-separated. We observe that there is a substantial increase in scores, both quantitatively and qualitatively (Fig. \ref{fig:poc-scores} and Fig. \ref{fig:poc-scores-for-AT}). Note that we maintain the original tokenizer for all experiments and add separation characters (i.e., whitespace or @) to the input text before tokenization. Although there is overlap between the resultant word-piece tokens and true morphemes, it is incidental and typically low (with {\tt GPT-4}, we observed only a 0.62 out of 1 Jaccard similarity between the morphemes and the word-pieces).

The results thus show that all models that were performing above 15\% on the subset solved more problems accurately after splitting. While {\tt Llama2-7B}, {\tt 13B}, and {\tt Llama3-8B} were mostly unaffected by the splits, other models like {\tt Llama3-70B} (Tab. \ref{tab:poc_qwise_gains}) reached 50\% accuracy, solving some puzzles better than the baseline scores of much bigger models. The substantial score improvements after morphological splitting demonstrate that morphology was indeed the primary bottleneck, while the remaining performance gaps in some puzzles reveal the contribution of non-linguistic reasoning components.

\begin{wrapfigure}[20]{}{0.45\textwidth}
    \centering
    \vspace{-0.5cm}
    \includegraphics[width=\linewidth]{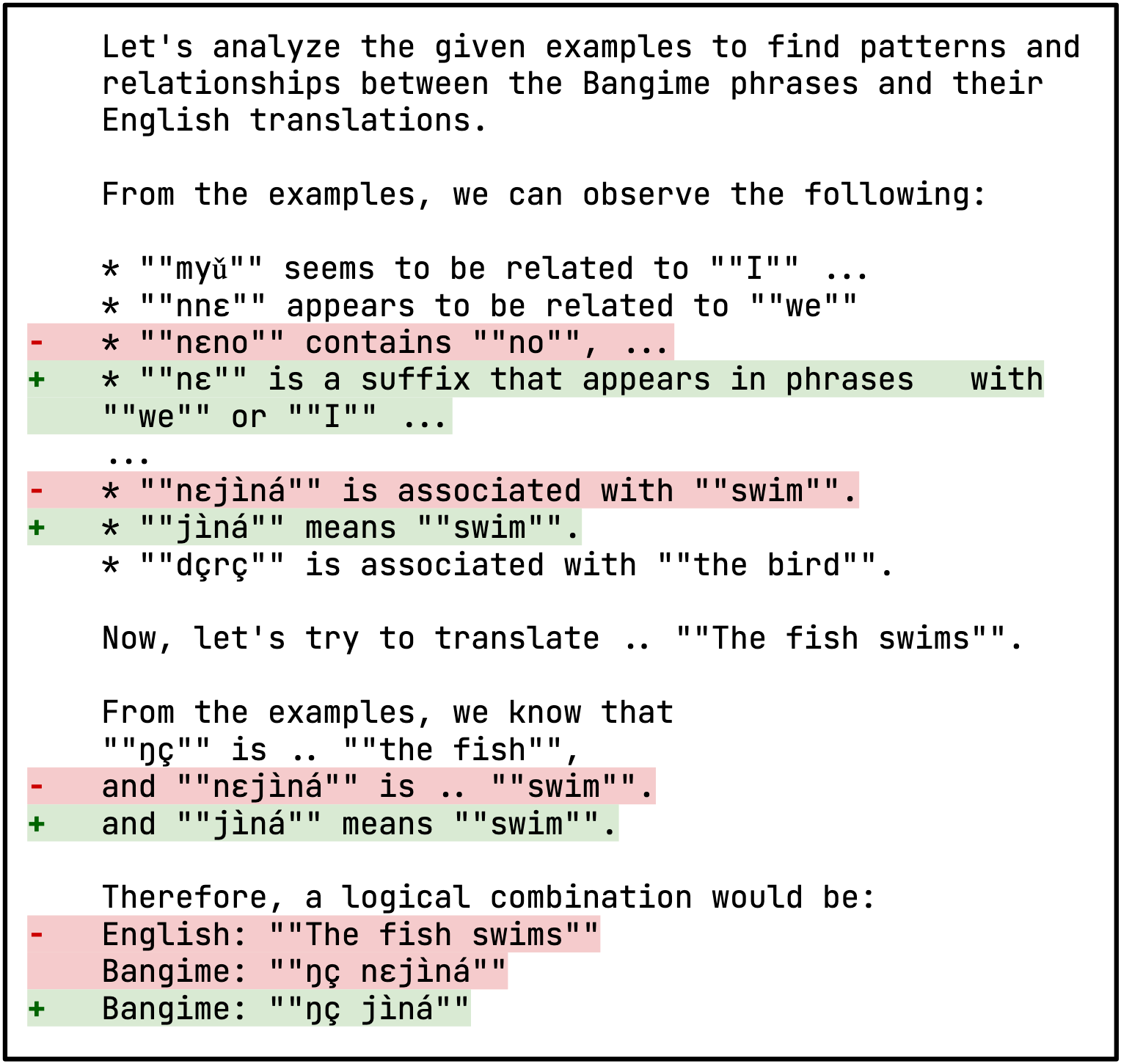}
    \caption{{\tt Llama3-70B} response change (\highlight{F4CCCC}{{\tt before}} to \highlight{D9EAD3}{{\tt after}}) recognises the person affix after splitting morphemes and reaches the correct final answer.}
    \vspace{-0.2cm}
    \label{fig:bg-ex}
    \vspace{-0.5cm}
\end{wrapfigure}

Furthermore, separating morpheme boundaries by a whitespace or by `@' did not lead to very different results. Fig. \ref{fig:bg-ex} shows {\tt Llama3-70B}'s response change after split on an extended version of the Bangime puzzle from Fig. \ref{fig:corr-plot}. The response was formatted and worded before and after splitting in exactly the same way by the LLM (we have added ellipses for simpler presentation), except that after marking the morpheme boundaries, it identified the right morphemes and their translations after splitting.

Finally, we also find that LLMs could not solve some puzzles with morphemes that packed a lot of semantics (e.g., kinship in Lardil) or puzzles with a lot of morphemes (e.g., Ayutla Mixe -- see Fig. \ref{fig:subfig2}).

However, for most cases (see Tab. \ref{tab:poc_qwise_gains_full}), we observe that just splitting a set of LRL or unseen language data into morphemes can lead to better translations.

\section{Discussion \& Related Work}

As interest grows in understanding how LLMs' processing of language aligns with linguistic theory \citep{futrell2025linguistics}, researchers find that LM abilities pertain to specific linguistic features rather than languages in the context of multilingual tasks \citep{acs2024morphosyntactic}. The \textsc{unveiLing} method probes LLMs in a similar fashion. We find that breaking LO puzzles down into linguistic attributes to study LLMs is reliable (i.e., it confirms some existing findings from an intrinsic perspective of linguistic features in the puzzle instead of Likert-based difficulty ratings \citep{chi-etal-2024-modeling}) and can be expanded (i.e., it allows us to unpack more about LLM performance).

\paragraph{Data:} We show that data constraint adds to a puzzle's difficulty for LLMs by annotating redundancy for each feature in a puzzle. Other research shows extrinsically that puzzles are made easier for LLMs by adding natural language data across linguistic abstractions \citep{zhu2025evaluating}, prompting models to generate data from a typologically close language \citep{ramji2024inductive}, or by augmenting data through character replacements \citep{watanabeexploring}.\footnote{For most of the work cited here, the data creation strategy is either not fully explained, or synthetic data is either linguistically infeasible or not validated by native speakers.}

\paragraph{English bias:} We show a feature-level typological bias towards linguistic features in English. Past research in LM performance on LO puzzles and results from Sec. \ref{sec:exp} arrive at this conclusion extrinsically by showing that there is also translation directionality bias towards English, where all models showed that LLMs are better at translating into English as compared to translating into an LRL. Outside of LO puzzles, in a cross-linguistic generalization ability probing of LLMs through a Wug Test \citep{anh-etal-2024-morphology}, authors found that most LLMs do not add affixes to nonce words, add them incorrectly, or even use English affixes in another lower-resourced language and context. 

Interestingly, \citet{warstadt-etal-2020-blimp-benchmark} find LMs to have concentrated robust knowledge of specific domains of grammar in English (based on supervised training on acceptability data of minimal pairs). They note that \textit{``however, these and related studies cover a limited set of phenomena, excluding ellipsis, quantification, and countless others. This is likely due to the labour-intensive nature of collecting such targeted minimal pairs''}. We believe that LO puzzles are by design positioned as a neat set of minimal pairs in extremely low-resourced languages, addressing this problem.

\paragraph{Morphology \& Tokenisation:} While our analyses show a significant correlation between poorer LLM performance and an increasing number of morphological features in an LO puzzle (based on an unseen natural language), most other LO research confirms this very superficially. Upon making morpheme boundaries explicit, we found that LLMs performed better generally. This implies a need for a deeper focus on revamping tokenisation approaches by either researching ways for tokenisers to adapt to language-specific contexts or using tokeniser-free approaches.

Much like our experiments, \citet{ismayilzada2024evaluating} investigate the morph generalization abilities of LLMs through the lens of compositionality and find a major gap in their performance as
compared to humans (in agglutinative languages). \citet{weissweiler2023counting} also apply a Wug test in typologically diverse languages (English, German, Tamil, and Turkish) to arrive at the same conclusion.

\citet{haslett2025tokenization} finds that this is because LLMs do not directly perceive units like affixes and semantic radicals. Here, subword tokens are determined by frequency, not semantics, and end up obscuring meaningful units. Larger vocabularies of longer units hide sub-morphemic units from LLMs. 

However, the tokenisation problem has been recently proven to be NP-complete \citep{whittington2024tokenisation}, and various new solutions, such as Greedy tokenisation \citep{lim2025partition}, Byte-free tokenisation (where dedicated local transformers identify dynamically sized patches of bytes) \citep{pagnoni2024byte}, or trigram-based hashing of vocabulary \citep{deiseroth-etal-2024-free}, might help with the morphology gap.

Another line of relevant work is to use language-specific tokenisers. \citet{dang2024tokenization, bayram2025tokenization} find that subword-based and morphologically informed tokenization works better for Turkish, whereas character-level tokenization benefits Hindi morphology. New methods like MorphBPE (BPE with an additional step of blocking pair-merging apart from morpheme boundaries) \citep{asgari2025morphbpe, lerner2025unlike} showed faster convergence in training and more alignment with morphologies of languages with different richness \citep{team2025fanar}.

Consequently, there is a need for more intrinsic approaches to evaluation  of model abilities in morphological processing. \citet{ali2024tokenizer} show that low fertility scores are a necessary but not sufficient criterion for this. \citet{arnett2024language} also claim that byte-premiums and scaling data need to be factored in with morphology-tokenisation alignment differences to explain performance degradation in LRLs. Based on our analyses of LLMs, we believe that designing translation puzzles in unseen languages (like LO puzzles), controlled for specific and rare morphological features, is a starting point for that.

\section{Conclusions \& Future Work}

We presented a linguistic feature-based breakdown of Linguistics Olympiad style puzzles and a correlation study between various features of linguistics puzzles and the performance of LLMs on these puzzles. Our analyses show that LLMs are better at solving puzzles with more English-like features, struggle to solve puzzles with more features that have constrained data to uncover them, and puzzles that have a higher number of morphological features in low-resource and unseen languages. This shows that LLMs can be studied in more depth in terms of their linguistic reasoning and puzzle-solving abilities and weaknesses in a Linguistics Olympiad-style setting.  

We further probed models by making morpheme boundaries explicit to unveil that LLMs performed better with this small change in the puzzles. This motivates a deeper look into tokenisation techniques used in multi-lingual settings, specifically for extremely low-resource languages or unseen data.

We believe that a linguistic feature-based analysis of LO puzzles can provide a better insight into how LLMs unpack new language data, and that we could design more targeted LO-style puzzles to unveil specific aspects. Other analyses could include addressing different kinds of morphology, e.g. inflectional as well as derivational, to better understand the observed effects. We could also explore how different prompting strategies change LLM responses and their consistency in solving LO puzzles \citep{lin-etal-2023-solving}.

Finally, we hope that this work may also inform other tasks such as studying multi-modal LLM capabilities in deciphering low-resource languages' orthographies \citep{shih2025reasoning} or in making educational language learning apps \citep{vasselli2024applying} annotated with hand-crafted features to bolster LLM abilities in modelling low resource languages.

% https://x.com/eduardosg_ai/status/1837192696353673593 o1 doesnt show outstanding improvement over 4o and Claude

% \section*{TOBEDONE Acknowledgments}
% Msc students and more

\section*{Ethics Statement}
With regard to the datasets used in our work, we comply with all available licenses and/or have sought authors' explicit permission to use their datasets. We have made minimal changes and will publish only mappable annotations and a list of fixes for full reproducibility. The annotators volunteered for the task and are co-authors of the paper. We foresee no other major ethical concerns that could arise from this work or the data annotated.

\section*{Reproducibility Statement}
All code and data are publicly released at \url{https://github.com/mukundc2k/unveiling}, ensuring that our experiments can be independently reproduced. To guarantee consistency, all experiments involving LLMs were performed with a temperature setting of 0. Detailed information about the specific model variants used and the complete querying specifications can be found in App. \ref{app:querying-details}. We also note that the contamination check performed in this study does not rule out the possibility that models released after the study may have been trained on the publicly available subset of puzzles used herein.

\bibliography{colm2025_conference}
\bibliographystyle{colm2025_conference}

\newpage

\appendix

% \section{Example Appendix}
% \label{sec:appendix}

\section{Querying Details}
\label{app:querying-details}

\subsection{Models}

Tab. \ref{tab:llm-versions} presents the corresponding model versions for the LLMs used in our experiments and analyses. All \texttt{GPT-*} models were accessed from a dedicated OpenAI API and all other models were imported from HuggingFace \citep{wolf2020huggingfacestransformersstateoftheartnatural} and loaded and inferred using vLLM.

\subsection{Computational Resources}

We query all models with a temperature of $0$ and a max output token length of 1024 for each problem response. The exact prompt templates used for querying are as follows:

\begin{table}[h]
\centering
\resizebox{0.5\linewidth}{!}{%
\begin{tabular}{@{}ll@{}}
\toprule
\textbf{LLM} & \textbf{Model Version} \\ \midrule
GPT-4 & \texttt{gpt-4-0613} \\ \midrule
GPT-4o & \texttt{gpt-4-turbo-2024-04-09} \\ \midrule
GPT-3.5-Turbo & \texttt{gpt-3.5-turbo-0125} \\ \midrule
Llama3.1-405B & \texttt{meta-llama/Llama-3.1-405B-Instruct} \\ \midrule
Llama3-70B & \texttt{meta-llama/Meta-Llama-3-70B-Instruct} \\ \midrule
Llama3-8B & \texttt{meta-llama/Meta-Llama-3-8B-Instruct} \\ \midrule
Llama2-70B & \texttt{meta-llama/Llama-2-70b-chat-hf} \\ \midrule
Llama2-13B & \texttt{meta-llama/Llama-2-13b-chat-hf} \\ \midrule
Llama2-7B & \texttt{meta-llama/Llama-2-7b-chat-hf} \\ \midrule
Mixtral-7x8B & \texttt{mistralai/Mixtral-8x7B-Instruct-v0.1} \\ \bottomrule
\end{tabular}%
}
\caption{Model versions for the LLMs used in our experiments and analyses.}
\label{tab:llm-versions}
\end{table}

\subsection{Prompts}
In this section, we present the various prompts used in our experiments and analyses.

\begin{figure}[!]
    \centering
    \includegraphics[width=0.5\linewidth]{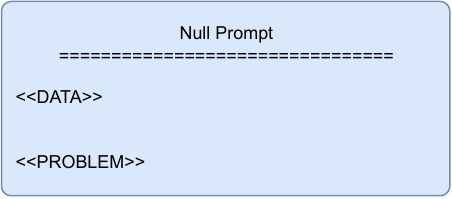}
    % \vspace{-0.4cm}
    \caption{Prompt template for \texttt{Null} prompt}
    % \vspace{-0.5cm}
    \label{fig:prompt-null}
\end{figure}

\begin{figure}[!]
    \centering
    \includegraphics[width=0.5\linewidth]{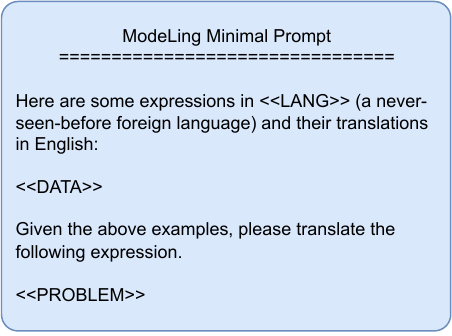}
    % \vspace{-0.4cm}
    \caption{Prompt template for \texttt{ModeLing Minimal} prompt}
    % \vspace{-0.5cm}
    \label{fig:prompt-m_minimal}
\end{figure}

\begin{figure}[!]
    \centering
    \includegraphics[width=0.5\linewidth]{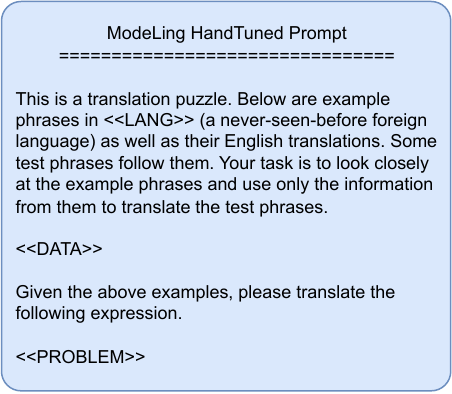}
    % \vspace{-0.4cm}
    \caption{Prompt template for \texttt{ModeLing HandTuned} prompt}
    % \vspace{-0.5cm}
    \label{fig:prompt-m_handtuned}
\end{figure}

\begin{figure}[!]
    \centering
    \includegraphics[width=0.5\linewidth]{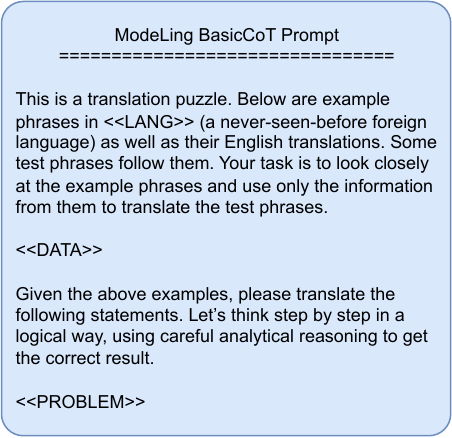}
    % \vspace{-0.4cm}
    \caption{Prompt template for \texttt{ModeLing BasicCot} prompt}
    % \vspace{-0.5cm}
    \label{fig:prompt-m_basiccot}
\end{figure}

\begin{figure}[!]
    \centering
    \includegraphics[width=0.5\linewidth]{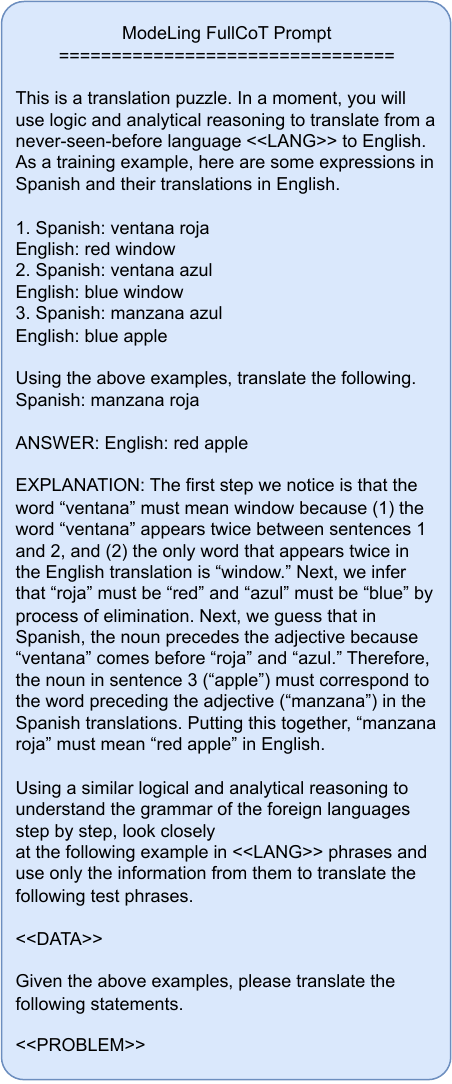}
    % \vspace{-0.4cm}
    \caption{Prompt template for \texttt{ModeLing FullCoT} prompt}
    % \vspace{-0.5cm}
    \label{fig:prompt-m_fullcot}
\end{figure}

\begin{figure}[!]
    \centering
    \includegraphics[width=0.5\linewidth]{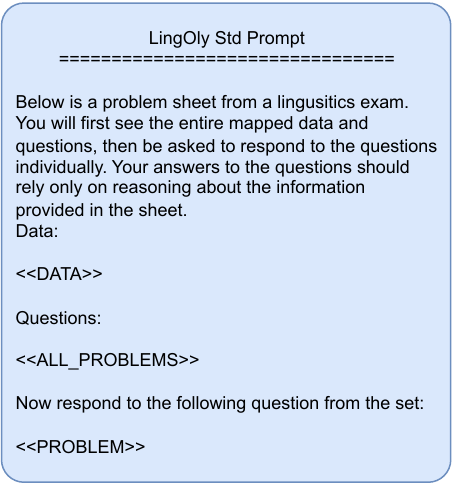}
    % \vspace{-0.4cm}
    \caption{Prompt template for \texttt{LingOly} prompt}
    % \vspace{-0.5cm}
    \label{fig:prompt-l_std}
\end{figure}

\begin{figure}[!]
    \centering
    \includegraphics[width=0.5\linewidth]{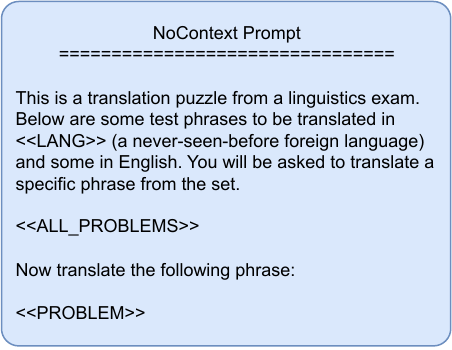}
    % \vspace{-0.4cm}
    \caption{Prompt template for \texttt{NoContext} prompt}
    % \vspace{-0.5cm}
    \label{fig:prompt-nocontext}
\end{figure}

\begin{enumerate}
    \item Null Prompt (Only data from the puzzle and one problem at a time, without any explicit instructions.) -- See Figure \ref{fig:prompt-null}.
    \item \textsc{modeLing} Minimal Prompt \citep{chi-etal-2024-modeling} -- See Figure \ref{fig:prompt-m_minimal}.
    \item \textsc{modeLing} Hand-Tuned Prompt \citep{chi-etal-2024-modeling} -- See Figure \ref{fig:prompt-m_handtuned}.
    \item \textsc{modeLing} Basic CoT Prompt \citep{chi-etal-2024-modeling} -- See Figure \ref{fig:prompt-m_basiccot}.
    \item \textsc{modeLing} Full CoT Prompt \citep{chi-etal-2024-modeling} -- See Figure \ref{fig:prompt-m_fullcot}.
    \item \textsc{LingOly} Std Prompt \citep{bean2024lingoly} -- See Figure \ref{fig:prompt-l_std}.
    \item No Context Prompt (Context pairs from data in the puzzle withheld for detecting contamination.) -- See Figure \ref{fig:prompt-nocontext}.
\end{enumerate}

\begin{figure}[!]
    \centering
    \includegraphics[width=0.5\linewidth]{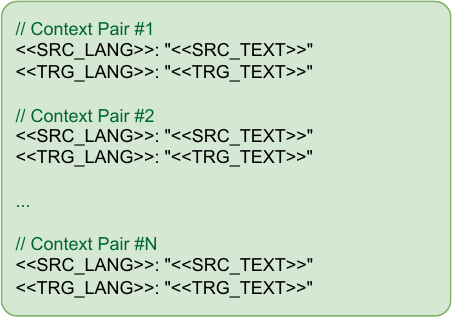}
    % \vspace{-0.4cm}
    \caption{Prompt template for \texttt{<<DATA>>} span. Lines starting with \texttt{"//"} are comments only for reference and not part of the actual span.}
    % \vspace{-0.5cm}
    \label{fig:prompt-data}
\end{figure}

\begin{figure}[!]
    \centering
    \includegraphics[width=0.5\linewidth]{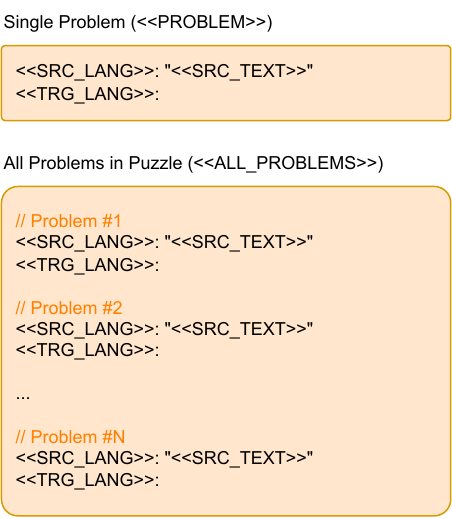}
    % \vspace{-0.4cm}
    \caption{Prompt template for \texttt{"<<PROBLEM>>"} and \texttt{"<<ALL\_PROBLEMS>>"} spans. Lines starting with \texttt{"//"} are comments only for reference and not part of the actual span.}
    % \vspace{-0.5cm}
    \label{fig:prompt-problem}
\end{figure}

In each of the below prompt templates, the span \texttt{"<<DATA>>"} corresponds to the translation pairs of the source language to the target language provided as data for a puzzle. This is presented in Figure \ref{fig:prompt-data}. The exact layout of an individual problem (\texttt{"<<PROBLEM>>"}) and all the problems (\texttt{"<<ALL\_PROBLEMS>>"}) in a puzzle are presented in Figure \ref{fig:prompt-problem}.

\newpage

\section{\textsc{modeLing} corrections}
\label{app:modeling-corrections}
Here we present a list of changes to the \textsc{modeLing} dataset that we made, along with our rationale. Each problem is referred to with a language name and a number if necessary.
\begin{itemize}
    \item \textbf{Ainu:} We removed ``='' symbols: they are used to help beginner readers understand morphology.\footnote{\href{https://unilang.org/course.php?res=58}{Ainu for beginners}} However, it is not clear, whether those symbols would help an LLM or confuse it.
    
    \item \textbf{Ayutla Mixe:} We changed ``Ëjts yë'ë nexpy'' in answers to ``Ëjts yë'ë ntunpy'' as a translation for ``I work for him''. ``Tun'' is the root for ``work'', while ``ex'' is the one for ``see''. Similarly, we changed ``Yë' maxu'unk yë' uk yexpy.'' to ``Yë' maxu'unk yë' uk yex\textbf{yejt}py.'' in the solutions for consistency.\footnote{\href{https://www.proquest.com/docview/305085835}{A reference grammar for Ayutla Mixe}}
    
    \item \textbf{Bangime 4:} We changed ``nnε dçrç nεnonnε'' in answers to ``nnε dçrç nεno'' as translation to ``we hear the bird'', because the  ``nnε'' suffix is not present in the examples in the question or even in the reference.\footnote{\href{http://www.rogerblench.info/Language/Isolates/Bangi\%20me\%20\%20paper\%20for\%20MT.pdf}{Bangime reference}}
    
    \item \textbf{Bangime 7:} We changed ``bad food'' in the questions to ``bad land'' (answer: ``gwyε begaũ'') because ``bad food'' is already in the given parallel sentences. We also changed ``big house'' to ``big food'' (answer: ``jyεkε kanyoro'') to maintain the same number for each noun in the data and questions.

    \item \textbf{Dogon}: We changed ``ú íló'' in solutions to ``ú ílò'' because the High-High (HH) pattern was not used anywhere in the examples (even in Dogon 2). The rule, as we understand it, is that the possessed noun tones are LL in the examples with non-pronominal noun possessors and HL in the pronominal noun possessors. This can be cross-checked in section 6.2.1.1 (pg. 141-143) of the reference.\footnote{\href{https://deepblue.lib.umich.edu/bitstream/handle/2027.42/123064/A?sequence=4}{A Grammar of Toro Tegu (Dogon) Tabi mountain dialect}}  

    \item \textbf{Kalam:} We changed the question ``kapkap yib ag'' to ``kamiket yib ag'' and the corresponding answer from ``speak very slowly'' to ``speak very stealthily'' as the word/root for ``slowly'' is not in the data.

    \item \textbf{Kutenai:} We replaced ``stomach'' in the questions with ``belly'' and ``wum'' in answers with ``wumnana''. The original question requires inferring from ``wumnana - belly'' that belly is a ``small stomach'', however official definitions state that ``belly'' is the general area of the stomach and adjuncts. In addition, ``nana - small'' is being tested in another question.

    \item \textbf{Mapudungan 1:} We changed ``pichi mansun'' to ``pichi pafu'' (answer: ``small turkey'') as the original phrase was translated already in the data.

    \item \textbf{Mixtepec Zapotec 2:} We changed ``zhàb ròsâd'' (pink clothing) to ``yâg ròsâd'' (pink stick) for the same reason as in the previous task.

    \item \textbf{Ngadha 1:} We changed ``big leaf'' in examples to ``big fruit'' as there is not enough data in the problem to establish that ``fruit'' (``li’e'') and ``leaf'' are the same (if they are) in the answers.

\end{itemize}

\section{Benchmarking details}
\label{app:bench}

We report the average exact match accuracy of each model-prompt combination in Tab. \ref{tab:model-prompt-acc} across our dataset (O = M+L) and across the M \& L subsets. 

Here is an overview of our benchmarking experiments, which show results in line with previous literature (Tab. \ref{tab:datasets-summary} and BIG-Bench \citep{srivastava2023beyond}), demonstrating that closed-source, larger models are generally better solvers, and puzzles set in low-resource languages are generally more difficult to solve.

\paragraph{Along Models and Prompts} We find that GPT-4o performs the best, and while {\tt GPT-4} scores slightly lower, it shows more consistency across prompts than other models. In general, larger models within a model family perform better. However, for suitable prompts, relatively smaller models -- {\tt Llama3-8B} and {\tt Mixtral-8x7B} -- outperform larger models like {\tt Llama2-70B} and {\tt GPT-3.5-Turbo}. This aligns with \citet{waldis2024holmes}'s observation that, while parameter size generally correlates with linguistic competence, specific architecture and tuning parameters can impact models' capabilities, particularly in the domains of morphology and syntax.

Consistent with the results of \citet{sanchez2024linguini}, who find that OpenAI models work better with fewer shot prompting, we also find that {\tt GPT-4o} and {\tt Mixtral-8x7B} perform better with more concise prompts (i.e., Null and Minimal).

Full CoT provides models with a descriptive solution to a short linguistic puzzle in Spanish on noun phrase construction and proceeds to instruct the models to solve the actual linguistic puzzle similarly. Thus, it is possible that many of the models perform well with the Full CoT prompt as it provides a short demonstration in a known language.

\paragraph{Along Dataset Sources} All models score higher on M puzzles than on L puzzles. As described earlier, this is because M puzzles are made in the style of LO puzzles but involve a lower number of linguistic features per question and have more data per feature to uncover it in the puzzle.

\begin{figure}[h]
    \centering
    \begin{subfigure}[b]{0.7\textwidth}
        \centering
        \includegraphics[width=\linewidth]{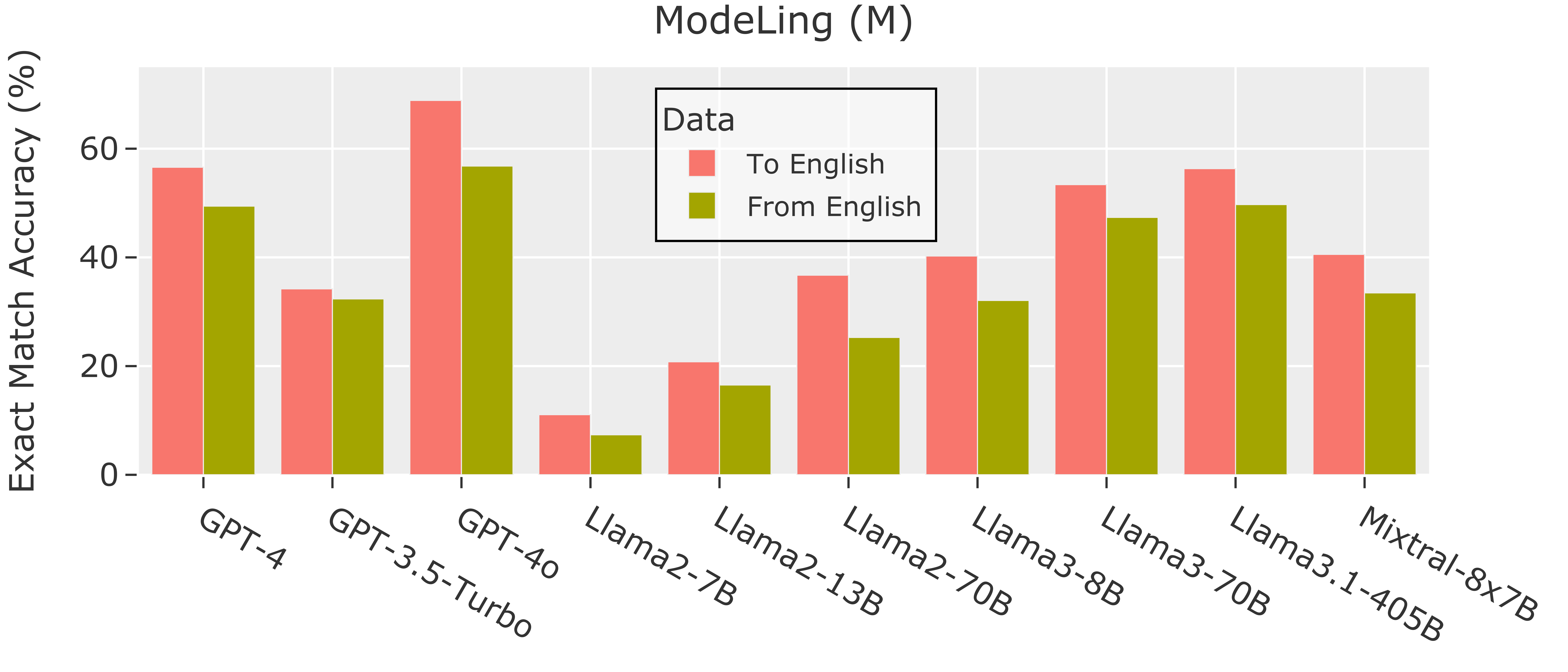}
        % \label{fig:sub1}
    \end{subfigure}
    % \hfill
    \vspace{-6.5mm}
    
    \begin{subfigure}[b]{0.7\textwidth}
        \centering
        \includegraphics[width=\linewidth]{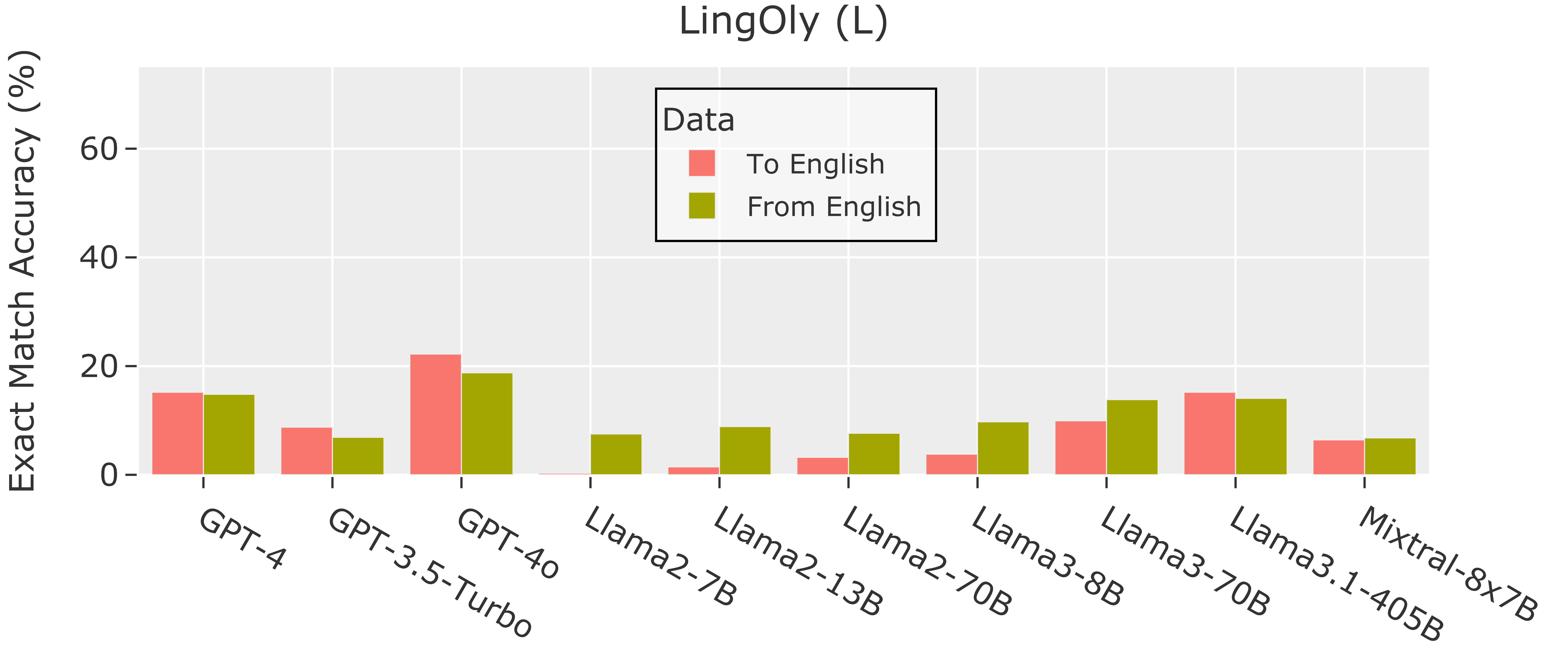}
        % \label{fig:sub2}
    \end{subfigure}
    % \vspace{-4mm}
    \caption{Models scores across translation directions.}
    \label{fig:translation_direction}
\end{figure}

\paragraph{Along Translation Direction} Consistent with \citet{csahin2020puzzling}'s findings for LLMs before 2021, we find that most models also score better when translating from the low-resource language {\em to} English compared to when translating {\em from} English (see Figure \ref{fig:translation_direction}). This difference is most pronounced in {\tt GPT-4o}'s results on M puzzles.

\subsection{Extended Results for DeepSeek-R1}
\label{app:subsec-r1-scores}
Table \ref{tab:r1-scores} presents the exact match accuracy on the entire evaluation set (O) for {\tt DeepSeek-R1}, which has been tuned for Inference-Time Compute \citep{deepseekai2025deepseekr1incentivizingreasoningcapability}. We compare these scores with the best scores for each prompt setting as shown in Table \ref{tab:model-prompt-acc}. We observe that R1 demonstrates superior performance while maintaining our observed correlations between model performance and morphological features (Pearson correlation = -0.74, p-value $<$ 0.001). To better determine the impact of ITC on this significant gain, future work must extend our study to a wider set of ITC models.

\begin{table}[]
\centering
\resizebox{0.5\columnwidth}{!}{%
\begin{tabular}{@{}lrrc@{}}
\toprule
\multirow{2}{*}{\textbf{Model}} & \multicolumn{2}{c}{\textbf{Average Accuracy}} & \multirow{2}{*}{\textbf{t}} \\ \cmidrule(lr){2-3}
 & \textbf{M} & \textbf{L} &  \\ \midrule
{\tt GPT-4} & 56.1 & 13.5 & 12.55*** \\
{\tt GPT-3.5-Turbo} & 35.1 & 6.6 & 9.99*** \\
{\tt GPT-4o} & 65.7 & 18.6 & 15.77*** \\
{\tt Llama2-7B} & 9.3 & 4.1 & 3.45** \\
{\tt Llama2-13B} & 19.4 & 5.1 & 6.33*** \\
{\tt Llama2-70B} & 31.9 & 5.3 & 10.77*** \\
{\tt Llama3-8B} & 38.4 & 6.6 & 11.05*** \\
{\tt Llama3-70B} & 53.2 & 11.2 & 13.03*** \\
{\tt Llama3.1-405B} & 55.3 & 14.1 & 11.93*** \\
{\tt Mixtral-8x7B} & 39.1 & 5.8 & 12.13*** \\ \bottomrule
\end{tabular}%
}
\caption{Average accuracy on M puzzles is significantly higher than L puzzles for all models. $t$-statistics marked with `***' indicates a $p$-value $\leq .001.$}
\label{tab:m_and_d_are_diff}
\end{table}

% (TODO ADITYA: Add to the legend that ModeLing is (M) and LongOly is (L))

\begin{table*}[th!]
\centering
\resizebox{\textwidth}{!}{%
\begin{tabular}{@{}lcccccccccccccccccc|ccc@{}}
\toprule[1.5pt]
 & \multicolumn{3}{c}{\textbf{Null Prompt}} & \multicolumn{3}{c}{\textbf{ModeLing Minimal}} & \multicolumn{3}{c}{\textbf{ModeLing Hand-Tuned}} & \multicolumn{3}{c}{\textbf{ModeLing Basic CoT}} & \multicolumn{3}{c}{\textbf{ModeLing Full CoT}} & \multicolumn{3}{c|}{\textbf{\textsc{LingOly} Std}} & \multicolumn{3}{c}{\textit{Model-Average}} \\ \cmidrule(l){2-22} 
\multirow{-2}{*}{\textbf{Model}} & \cellcolor[HTML]{EFEFEF}O & M & L & \cellcolor[HTML]{EFEFEF}O & M & L & \cellcolor[HTML]{EFEFEF}O & M & L & \cellcolor[HTML]{EFEFEF}O & M & L & \cellcolor[HTML]{EFEFEF}O & M & L & \cellcolor[HTML]{EFEFEF}O & M & L & \cellcolor[HTML]{EFEFEF}O & M & L \\ \midrule[1pt]
GPT-4 & \cellcolor[HTML]{EFEFEF}41.7 & \underline{58.2} & \textbf{12.3} & \cellcolor[HTML]{EFEFEF}41.3 & 58.1 & 11.4 & \cellcolor[HTML]{EFEFEF}40.2 & 54.3 & 15.0 & \cellcolor[HTML]{EFEFEF}41.1 & 55.7 & 15.2 & \cellcolor[HTML]{EFEFEF}42.0 & 56.9 & 15.3 & \cellcolor[HTML]{EFEFEF}38.4 & 53.6 & 11.4 & \cellcolor[HTML]{EFEFEF}40.8 & 56.1 & 13.5 \\
GPT-3.5-Turbo & \cellcolor[HTML]{EFEFEF}5.2 & 6.4 & 3.1 & \cellcolor[HTML]{EFEFEF}25.8 & 37.6 & 4.8 & \cellcolor[HTML]{EFEFEF}33.4 & \underline{47.3} & 8.6 & \cellcolor[HTML]{EFEFEF}25.3 & 34.6 & 8.5 & \cellcolor[HTML]{EFEFEF}32.8 & 47.2 & 7.3 & \cellcolor[HTML]{EFEFEF}26.5 & 37.4 & 7.1 & \cellcolor[HTML]{EFEFEF}24.8 & 35.1 & 6.6 \\
GPT-4o & \cellcolor[HTML]{EFEFEF}\textbf{47.4} & \textbf{67.1} & 12.2 & \cellcolor[HTML]{EFEFEF}\textbf{53.1} & \textbf{\underline{71.7}} & \textbf{19.8} & \cellcolor[HTML]{EFEFEF}\textbf{49.2} & 66.0 & \textbf{19.4} & \cellcolor[HTML]{EFEFEF}47.6 & 60.9 & \textbf{23.8} & \cellcolor[HTML]{EFEFEF}47.0 & 62.9 & \textbf{18.5} & \cellcolor[HTML]{EFEFEF}\textbf{48.3} & \textbf{65.4} & \textbf{17.8} & \cellcolor[HTML]{EFEFEF}\textbf{48.8} & \textbf{65.7} & \textbf{18.6} \\ \midrule
Llama2-7B & \cellcolor[HTML]{EFEFEF}7.0 & 7.0 & 7.0 & \cellcolor[HTML]{EFEFEF}7.0 & 9.3 & 2.9 & \cellcolor[HTML]{EFEFEF}8.3 & 11.5 & 2.6 & \cellcolor[HTML]{EFEFEF}8.9 & \underline{11.8} & 3.7 & \cellcolor[HTML]{EFEFEF}8.7 & 10.2 & 5.9 & \cellcolor[HTML]{EFEFEF}4.7 & 5.8 & 2.7 & \cellcolor[HTML]{EFEFEF}7.4 & 9.3 & 4.1 \\
Llama2-13B & \cellcolor[HTML]{EFEFEF}8.6 & 8.8 & 8.1 & \cellcolor[HTML]{EFEFEF}14.4 & 19.8 & 4.7 & \cellcolor[HTML]{EFEFEF}20.1 & \underline{28.8} & 4.7 & \cellcolor[HTML]{EFEFEF}14.4 & 20.3 & 3.9 & \cellcolor[HTML]{EFEFEF}19.0 & 26.2 & 6.3 & \cellcolor[HTML]{EFEFEF}9.1 & 12.6 & 2.9 & \cellcolor[HTML]{EFEFEF}14.3 & 19.4 & 5.1 \\
Llama2-70B & \cellcolor[HTML]{EFEFEF}18.6 & 26.6 & 4.2 & \cellcolor[HTML]{EFEFEF}23.5 & 34.7 & 3.5 & \cellcolor[HTML]{EFEFEF}25.3 & 36.1 & 6.1 & \cellcolor[HTML]{EFEFEF}22.4 & 30.8 & 7.3 & \cellcolor[HTML]{EFEFEF}25.9 & \underline{36.2} & 7.5 & \cellcolor[HTML]{EFEFEF}18.6 & 27.1 & 3.4 & \cellcolor[HTML]{EFEFEF}22.4 & 31.9 & 5.3 \\ \midrule
Llama3-8B & \cellcolor[HTML]{EFEFEF}17.5 & 24.4 & 5.2 & \cellcolor[HTML]{EFEFEF}30.4 & \underline{43.3} & 7.4 & \cellcolor[HTML]{EFEFEF}29.4 & 42.8 & 5.4 & \cellcolor[HTML]{EFEFEF}27.8 & 38.3 & 9.1 & \cellcolor[HTML]{EFEFEF}30.2 & 42.8 & 7.6 & \cellcolor[HTML]{EFEFEF}26.4 & 38.5 & 5.0 & \cellcolor[HTML]{EFEFEF}26.9 & 38.4 & 6.6 \\
Llama3-70B & \cellcolor[HTML]{EFEFEF}34.3 & 50.8 & 5.0 & \cellcolor[HTML]{EFEFEF}37.0 & 52.4 & 9.6 & \cellcolor[HTML]{EFEFEF}37.8 & 50.3 & 15.3 & \cellcolor[HTML]{EFEFEF}40.5 & \underline{55.4} & 13.7 & \cellcolor[HTML]{EFEFEF}39.8 & \underline{55.4} & 12.2 & \cellcolor[HTML]{EFEFEF}39.3 & 54.7 & 11.7 & \cellcolor[HTML]{EFEFEF}38.1 & 53.2 & 11.2 \\ \midrule
Llama3.1-405B & \cellcolor[HTML]{EFEFEF}20.7 & 27.5 & 8.5 & \cellcolor[HTML]{EFEFEF}31.7 & 43.0 & 11.6 & \cellcolor[HTML]{EFEFEF}48.6 & \textbf{66.5} & 16.8 & \cellcolor[HTML]{EFEFEF}\textbf{48.7} & \textbf{64.1} & 21.2 & \cellcolor[HTML]{EFEFEF}\textbf{49.0} & \textbf{\underline{67.8}} & 15.6 & \cellcolor[HTML]{EFEFEF}44.3 & 63.0 & 10.8 & \cellcolor[HTML]{EFEFEF}40.5 & 55.3 & 14.1 \\ \midrule
Mixtral-8x7B & \cellcolor[HTML]{EFEFEF}23.9 & 35.5 & 3.3 & \cellcolor[HTML]{EFEFEF}32.5 & 45.3 & 9.8 & \cellcolor[HTML]{EFEFEF}31.8 & \underline{46.0} & 6.5 & \cellcolor[HTML]{EFEFEF}24.0 & 34.6 & 5.2 & \cellcolor[HTML]{EFEFEF}23.1 & 33.0 & 5.3 & \cellcolor[HTML]{EFEFEF}27.2 & 39.9 & 4.6 & \cellcolor[HTML]{EFEFEF}27.1 & 39.1 & 5.8 \\ \midrule[1pt]
\textit{Prompt Average} & \cellcolor[HTML]{EFEFEF}22.5 & 31.2 & 6.9 & \cellcolor[HTML]{EFEFEF}29.7 & 41.5 & 8.6 & \cellcolor[HTML]{EFEFEF}32.4 & \underline{45.0} & 10.1 & \cellcolor[HTML]{EFEFEF}30.1 & 40.7 & 11.2 & \cellcolor[HTML]{EFEFEF}31.7 & 43.9 & 10.2 & \cellcolor[HTML]{EFEFEF}28.3 & 39.8 & 7.7 & \cellcolor[HTML]{EFEFEF}29.1 & 40.3 & 9.1 \\ 
\bottomrule[1.5pt]
\end{tabular}%
}
\caption{Summary of an average exact match accuracy of each model-prompt combination across our dataset (O) and subsets of puzzles from \textsc{modeLing} (M) and \textsc{LingOly} (L). Models are divided by family and arranged by parameter size. \textbf{Bold} scores mark the best model for each prompt, \underline{underlined} scores mark the best prompts per model.}
\label{tab:model-prompt-acc}
\end{table*}

\begin{table}[]
\centering
\resizebox{0.4\textwidth}{!}{%
\begin{tabular}{@{}lcc@{}}
\toprule
\textbf{Prompt} & \multicolumn{1}{c}{\textbf{\begin{tabular}[c]{@{}c@{}}Previous\\ Best\end{tabular}}} & \multicolumn{1}{c}{\textbf{\begin{tabular}[c]{@{}c@{}}Score for \\ DeepSeek-R1\end{tabular}}} \\ \midrule
NullPrompt & 47.4 & 60.7 \\ \midrule
ModeLing Minimal & 53.1 & 61.4 \\ \midrule
ModeLing Hand-Tuned & 49.2 & 64.7 \\ \midrule
ModeLing Basic-CoT & 48.7 & 64.9 \\ \midrule
ModeLing Full-CoT & 49.0 & 60.5 \\ \midrule
LingOly Std & 48.3 & 61.9 \\ \bottomrule
\end{tabular}%
}
\caption{Extended results for {\tt DeepSeek-R1} compared against the previous best scores across both datasets (O) for each prompt setting (see Table \ref{tab:model-prompt-acc}).}
\label{tab:r1-scores}
\end{table}

% \section{Feature List and Mappings}
% \label{sec:app-feat_list}

% \begin{figure}
%     \centering
%     \includegraphics[width=\linewidth]{figures/lo_feature_crosstabs.png}
%     % \vspace{-0.4cm}
%     \caption{}
%     % \vspace{-0.5cm}
%     \label{fig:feature_crosstab}
% \end{figure}

\section{Linguistic Features, Definitions, and Classification}
\label{sec:app-feat_list}

See Tab. \ref{tab:feat-def}.
% WORKING ON THIS (DONE, PLS SANITY CHECK)

\begin{table*}[]
\centering
\renewcommand{\arraystretch}{1.3} % Adjust row height for readability
\rowcolors{2}{gray!10}{white} % Alternate row colors
\resizebox{\textwidth}{!}
{%
\begin{tabular}{|>{\centering\arraybackslash}p{2.5cm}|>{\centering\arraybackslash}p{3cm}|>{\centering\arraybackslash}p{3cm}|>{\centering\arraybackslash}p{4cm}|>{\centering\arraybackslash}p{2.5cm}|>{\centering\arraybackslash}p{4cm}|>{\centering\arraybackslash}p{4cm}|}
\hline
\rowcolor{gray!25} % Header row color
\textbf{WALS Feature} & \textbf{WALS Category} & \textbf{Broad category} & \textbf{Definition} & \textbf{Example Problem} & \textbf{Example Problem Citation} & \textbf{Other Research} \\
\hline
59 - Possessive Classification \textit{(alien)} & Nominal Syntax & Semantics & the possessed object could be either close to the possessor (kinship, body part terms, etc.) or not & M\_Abun & \href{https://core.ac.uk/download/pdf/160609395.pdf}{A Description of Abun: A West Papuan Language of Irian Jaya (5.3.1 and 5.3.2)} & \href{https://wals.info/chapter/59}{WALS - Poss. Class.} \\ 
\hline
57 - Position of Poss. Markers \textit{(poslinker)} & Nominal Syntax & Syntax & a possessive particle that  between two nouns & M\_Abun & \href{https://core.ac.uk/download/pdf/160609395.pdf}{A Description of Abun: A West Papuan Language of Irian Jaya (5.3.1 and 5.3.2)} & \href{https://www.degruyter.com/document/doi/10.1515/tlr-2014-0024/html?lang=en}{Linkers and agreement} \href{https://wals.info/chapter/57}{WALS - Pos. of Poss. Markers} \\ 
\hline
13 - Tone \textit{(tone)} & Phonology & Phonology & pitch pattern grammar & M\_Dogon & \href{https://deepblue.lib.umich.edu/bitstream/handle/2027.42/123064/A?sequence=4}{A Grammar of Dogon} & \href{https://wals.info/chapter/13}{WALS - Tone} \\ 
\hline
101 - Expression of Pronominal Subjects \textit{(pron)} & Simple Clauses & Syntax & phonological rules or Lexical representation changed by pronoun role & M\_Dogon & \href{https://deepblue.lib.umich.edu/bitstream/handle/2027.42/123064/A?sequence=4}{A Grammar of Dogon (6.2.1.1)} & \href{https://wals.info/chapter/101}{WALS - Expr. of Pron. Subj.} \\ 
\hline
33 - Nominal Plurality \textit{(pl)} & Nominal Categories & Syntax & marker indicating plurality of noun & M\_Rapa\_Nui\_6 & \href{https://langsci-press.org/catalog/book/124}{A grammar of Rapa Nui - Paulus Kieviet 5.5} & \href{https://wals.info/chapter/33}{WALS - Nom. Pl.} \\ 
\hline
102 - Verbal Person Marking (\textit{transfix}) & Simple Clauses & Morphology & affix to mark transitiveness (agent+patient) on verb & M\_Ayutla\_Mixe & \href{https://www.proquest.com/openview/c03d979f7d2834985309564c6da4342f/1.pdf}{A reference grammar of Ayutla Mixe (Tukyo’m ayuujk) (6.3.3.1 and 8.2.4)} & \href{https://wals.info/chapter/102}{WALS - Verb. Pers. Marking} \\ 
\hline
102 - Verbal Person Marking \textit{(persfix)} & Simple Clauses & Morphology & affixes to mark person (only agent) on verb & M\_Ayutla\_Mixe & \href{https://www.proquest.com/openview/c03d979f7d2834985309564c6da4342f/1.pdf}{A reference grammar of Ayutla Mixe (Tukyo'm ayuujk) 8.2.1} & \href{https://wals.info/chapter/102}{WALS - Verb. Pers. Marking} \\ 
\hline
81-82 - Order of SOV, SV \textit{(sv/vso/vs/vos..)} & Word Order & Syntax & order of subject, object, and verb in a sentence. + orders for clauses without objects & M\_Ayutla\_Mixe & \href{https://www.proquest.com/openview/c03d979f7d2834985309564c6da4342f/1.pdf}{A reference grammar of Ayutla Mixe (Tukyo'm ayuujk) 9.3.1.2} & \href{https://wals.info/chapter/81}{WALS - Order of SOV} \href{https://wals.info/chapter/82}{WALS - Order of SV} \\ 
\hline
87 - Order of Adj., Noun \textit{(nounadj/adjnoun)} & Word Order & Syntax & order of noun and adjective in a noun phrase & M\_Bangime\_6 & \href{https://citeseerx.ist.psu.edu/document?repid=rep1\&type=pdf\&doi=ae0c0c51e742b10269f9c5c32954e22185883132}{Bangi me, a language of unknown affiliation 5.2} & \href{https://wals.info/feature/87}{WALS - Order of N-A} \\
\hline
NA - Diminutives \textit{(dim}) & Nominal Categories & Semantics & marker to express the little modifier to a noun / verb & M\_Kutenai & \href{https://en.wiktionary.org/wiki/Appendix:Kutenai_word_list}{Kutenai lexicon} & \href{https://www.researchgate.net/publication/372165276_Introduction_Diminutives_across_languages_theoretical_frameworks_and_linguistic_domains_httpslingaufnetlingbuzz007399}{Intro: Dim. across languages, theories ...} \textit{(WALS doesn't classify it)} \\
\hline
112 - Negative Morphemes \textit{(neg)} & Simple Clauses & Semantics & particle to express negation/absence of a noun & M\_Kutenai & \href{https://en.wiktionary.org/wiki/Appendix:Kutenai_word_list}{Kutenai lexicon} & \href{https://wals.info/chapter/112}{WALS - Neg. Morph.} \\
\hline
NA - Other Semantic Affixes \textit{(semfix)} & Nominal Categories & Morphology & affixes that add to the meaning to lexical categories & M\_Guugu\_Yimithir & \href{https://pure.mpg.de/rest/items/item_66622_3/component/file_66623/content}{Lang. \& Cognition: The Cog. Consequences of Spatial Description in Guugu Yimithirr pg. 4} & \href{https://www.academypublication.com/issues/past/tpls/vol03/05/04.pdf}{Prefixes of Spatiality in English: A Study in Cognitive Linguistics} \\
\hline
55 - Numeral Classifiers \textit{(cls)} & Nominal Categories & Semantics & classifier marker for nouns & M\_Ngadha\_2 & \href{https://core.ac.uk/download/pdf/160608296.pdf}{NGADHA TEXT TRADITION - Stephanus Djawana 2.2.3} & \href{https://pure.mpg.de/rest/items/item_59721_2/component/file_468401/content}{Nominal Classification} \href{https://wals.info/feature/55}{/ WALS - Num. Class.} \\
\hline
131 - Numeral Bases \textit{(num)} & Lexicon & Semantics & involves counting, numerals, and how positive whole number expressions are formed & M\_Ngadha\_2 & \href{https://core.ac.uk/download/pdf/160608296.pdf}{NGADHA TEXT TRADITION - Stephanus Djawana 3.5.1.3} & \href{http://www.lel.ed.ac.uk/~jim/numsys.pdf}{Numeral Systems} \href{https://wals.info/feature/131}{/ WALS - Num. Bases} \\
\hline
NA - Compounds \textit{comp} & Morphology & Semantics & compositionality: meanings of parts in an expression combine to convey something bigger & M\_Kalam & \href{https://www.researchgate.net/profile/Andrew-Pawley/publication/300841393_On_the_origins_of_serial_verb_constructions_in_Kalam/links/58d0fd8b4585158476f36662/On-the-origins-of-serial-verb-constructions-in-Kalam.pdf}{On the origins of serial verb constructions in Kalam} & \href{https://doi.org/10.1093/acrefore/9780199384655.013.563}{Attributive Compounds} \textit{(WALS doesn't classify this)}\\
\hline
125 - Purpose Clauses \textit{(svc)} & Complex Sentences & Syntax & serial verb construction: verbs combine in a single clause without any morphosyntactic marking of linking or subordination & M\_Kalam & \href{https://www.researchgate.net/profile/Andrew-Pawley/publication/300841393_On_the_origins_of_serial_verb_constructions_in_Kalam/links/58d0fd8b4585158476f36662/On-the-origins-of-serial-verb-constructions-in-Kalam.pdf}{On the origins of serial verb constructions in Kalam} & \href{https://www.annualreviews.org/docserver/fulltext/linguistics/7/1/annurev-linguistics-031920-115317.pdf?expires=1728064013\&id=id\&accname=guest\&checksum=86F69F32C158B8B068476D3EBD11A972}{Serial Verb Constructions} \href{https://wals.info/feature/125}{WALS - Purp. Clause.}  \\
\hline
24 - Possessive Noun Phrases \textit{(nounposs)} & Nominal Syntax & Syntax & order of noun and possessive marker in a noun phrase & M\_Seri\_4 & \href{https://www.sil.org/system/files/reapdata/10/62/88/106288372313712823757072453386621967679/G001_Sinopsis_sei.pdf}{Seri Grammar} & \href{https://wals.info/chapter/24}{WALS - Poss. NP} \\
\hline
85 - Order of Adposition and Noun Phrase \textit{(nounprep ..)} & Word Order & Syntax & order of noun and preposition in a prepositional phrase & M\_Seri\_2 & \href{https://www.sil.org/system/files/reapdata/10/62/88/106288372313712823757072453386621967679/G001_Sinopsis_sei.pdf}{Seri Grammar} & \href{https://wals.info/chapter/85}{WALS - Order of Adp. \& NP} \\
\hline
60 - Adjectival Clauses \textit{(nounadjc)} & Nominal Syntax & Syntax & order of noun and adjectival clause in a noun phrase being noun then adjectival clause & M\_Seri\_2 & \href{https://www.sil.org/system/files/reapdata/10/62/88/106288372313712823757072453386621967679/G001_Sinopsis_sei.pdf}{Seri Grammar 2.4} & \href{https://wals.info/chapter/60}{WALS - Gen., Adj. and Rel. Cls.} \\
\hline
24 - Possessive Noun Phrases \textit{(possfix)} & Nominal Syntax & Morphology & possessives marked as affix & M\_Totonac & \href{https://escholarship.org/uc/item/67p6q0xh}{The phonology and morphology of Filomeno Mata Totonac 3.3.2} & \href{https://wals.info/chapter/24}{WALS - Poss. Noun Phr.} \\
\hline
32 - Gender \textit{(genfix)} & Nominal Categories & Morphology & gender marking affixes & L\_Beja & \href{https://www.uklo.org/wp-content/uploads/2022/09/2013r2.3-Beja.pdf}{UKLO - Beja} & \href{https://wals.info/chapter/32}{WALS - Gender} \\
\hline

\end{tabular}%
}
\label{tab:feat-def-2}
\end{table*}

\begin{table*}[]
\centering
\renewcommand{\arraystretch}{1.3} % Adjust row height for readability
\rowcolors{1}{gray!10}{white} % Alternate row colors
\resizebox{\textwidth}{!}
{%
\begin{tabular}{|>{\centering\arraybackslash}p{2.5cm}|>{\centering\arraybackslash}p{3cm}|>{\centering\arraybackslash}p{3cm}|>{\centering\arraybackslash}p{4cm}|>{\centering\arraybackslash}p{2.5cm}|>{\centering\arraybackslash}p{4cm}|>{\centering\arraybackslash}p{4cm}|}
\hline
117 - Predicative Possession \textit{(poss)} & Simple Clauses & Syntax & possessive particles & L\_Coptic & \href{https://www.uklo.org/wp-content/uploads/2024/03/2024-R2_4-Coptic.pdf}{UKLO - Coptic} & \href{https://wals.info/chapter/117}{WALS - Pred. Poss.} \\
\hline
69 - Position of Tense-Aspect Affixes \textit{(tensefix)} & Verbal Categories & Morphology & tense marking affixes & L\_K’iche’ & \href{https://www.uklo.org/wp-content/uploads/2023/03/2023_R1_9-Kiche.pdf}{UKLO - K’iche’} & \href{https://wals.info/chapter/69}{WALS - Pos. of Tense-Asp. Aff.} \\
\hline
47 - Intensifiers \& Reflexive Pronouns \textit{(intens)} & Nominal Categories & Syntax & particle for intensifiers & L\_K’iche’ & \href{https://www.uklo.org/wp-content/uploads/2023/03/2023_R1_9-Kiche.pdf}{UKLO - K’iche’} & \href{https://www.cambridge.org/core/journals/english-today/article/abs/thats-proper-cool/C8D912AD2B00F207CD854455E27104A8}{``That's proper cool'': The emerging intensifier proper in British English} \href{https://wals.info/chapter/69}{/ WALS - Int. \& Refl. Pro.} \\
\hline
NA - Elision \textit{(eli)} & Morphology & Phonology & when a sound is elided during a fusion process, morphophonologically & L\_Kabyle & \href{https://www.uklo.org/wp-content/uploads/2022/05/2021_2-Kabyle.pdf}{UKLO - Kabyle} & \href{https://bpb-us-e1.wpmucdn.com/wp.nyu.edu/dist/e/1831/files/2015/04/LD_Phonetica_2006_offprint.pdf}{Schwa Elision in Fast Speech: Segmental Deletion or Gestural Overlap?} \textit{WALS doesn't classify this} \\
\hline
NA - Kinship order markers \textit{(moiety)} & Nominal Categories & Semantics & the concept of moiety encoded in affixes & L\_Lardil & \href{https://www.uklo.org/wp-content/uploads/2023/03/2023_R1_7-Lardil.zip}{UKLO - Lardil} & \href{https://www.degruyter.com/document/doi/10.1515/9783110279771.295/html}{7. Semantics of Australian Languages} \textit{WALS doesn't classify this} \\
\hline
34 - Nominal Plurality \textit{(numfix)} & Nominal Categories & Morphology & number of subject/object marking affixes & L\_Mayangna & \href{https://www.uklo.org/wp-content/uploads/2022/05/2018_R2_5-Mayangna.pdf}{UKLO - Mayangna} & \href{https://wals.info/chapter/34}{WALS - Occ. of Nom. Pl.} \\
\hline
49 - Number of Cases \textit{(case)} & Nominal Categories & Morphology & case marking & L\_Nhanda & \href{https://www.uklo.org/wp-content/uploads/2022/05/2016_9.-Nhanda.pdf}{UKLO - Nhanda} & \href{https://wals.info/chapter/49}{WALS - No. of Cases} \\
\hline
37 - Definite Articles \textit{(def)} & Nominal Categories & Morphology & definitiveness marking affixes & L\_Nhanda & \href{https://www.uklo.org/wp-content/uploads/2022/05/2016_9.-Nhanda.pdf}{UKLO - Nhanda} & \href{https://wals.info/chapter/37}{WALS - Def. Art. (the affixes bit)} \\
\hline
112 - Negative Morphemes \textit{(negfix)} & Simple Clauses & Morphology & negation marking affixes & L\_Tadaksahak & \href{https://www.uklo.org/wp-content/uploads/2022/09/2011r2.5-Tadaksahak.pdf}{UKLO - Tadaksahak} & \href{https://wals.info/chapter/112}{WALS - Neg. Morph.} \\
\hline
107 - Passive Constructions \textit{(voice)} & Simple Clauses & Syntax & active/passive voice having a morpho-syntactic impact & L\_Tadaksahak & \href{https://www.uklo.org/wp-content/uploads/2022/09/2011r2.5-Tadaksahak.pdf}{UKLO - Tadaksahak} & \href{https://wals.info/chapter/107}{WALS - Pass. Con.} \\
\hline
32 - Systems of Gender Assignment \textit{(anim)} & Nominal Categories & Semantics & whether noun has animacy or not & L\_Taos & \href{https://www.uklo.org/wp-content/uploads/2022/05/2022_R2_5_Taos.pdf}{UKLO - Taos} & \href{https://www.sciencedirect.com/science/article/pii/S0388000121000462}{Cog. anim. and its rel. to ling. anim.} \href{https://wals.info/chapter/32}{/ WALS - Sys. of Gen.} \\
\hline
27 - Reduplication \textit{(redup)} & Morphology & Morphology & grammatical partial/full reduplication of syllables & L\_Tawala & \href{https://www.uklo.org/wp-content/uploads/2022/05/2021_R2_5-Tawala.pdf}{UKLO - Tawala} & \href{https://wals.info/chapter/27}{WALS - Reduplication} \\
\hline
NA - Other Syntax Rules \textit{(miscsyn)} & Complex Sentences & Syntax & other syntactic rules like semantic placement of negation particles, or having an aspect marking particle at the end of each sentence & L\_Tawala & \href{https://www.uklo.org/wp-content/uploads/2022/05/2021_R2_5-Tawala.pdf}{UKLO - Tawala} &  \\
\hline
NA - Other Morphology Rules \textit{(miscfix)} & Morphology & Morphology & other affixes like "normalizer" & L\_Tseltal & \href{https://www.uklo.org/wp-content/uploads/2022/05/9_Adv_UKLO-2022-Tseltal__Complete-Script.pdf}{UKLO - Tseltal} &  \\
\hline
NA - Other Phonology Rules \textit{(miscphon)} & Phonology & Phonology & other phonological features like assimilation & L\_Waanyi & \href{https://www.uklo.org/wp-content/uploads/2022/09/2012.9-Waanyi.pdf}{UKLO - Waanyi} & \\
\hline
75 - Epistemic Possibility \textit{(modfix)} & Verbal Categories & Morphology & modality marking affixes & L\_Zou & \href{https://www.uklo.org/wp-content/uploads/2024/04/2024_R1_9-Zou.pdf}{UKLO - Zou} & \href{https://wals.info/chapter/75}{WALS - Epi. Poss.} \\
\hline
51 - Position of Case Affixes \textit{(prepfix)} & Nominal Categories & Morphology & prepositional affixes & L\_Zou & \href{https://www.uklo.org/wp-content/uploads/2024/04/2024_R1_9-Zou.pdf}{UKLO - Zou} & \href{https://wals.info/chapter/51}{WALS - Pos. of Case Aff. (prep clitics bit)} \\
\hline
116 - Polar Questions \textit{(quesfix)} & Simple Clauses & Morphology & interrogation affixes & L\_Zou & \href{https://www.uklo.org/wp-content/uploads/2024/04/2024_R1_9-Zou.pdf}{UKLO - Zou} & \href{https://wals.info/chapter/116}{WALS - Pol. Ques.} \\
\hline
95 - Order of Obj. \& Verb and Adp. \& NP \textit{(prepc)} & Word Order & Syntax & prepositional syntax with respect to the head clause & M\_Ainu & \href{https://www.researchgate.net/publication/233557551_Ainu_applicatives_in_typological_perspective}{(PDF) Ainu applicatives in typological perspective} & \href{https://wals.info/chapter/95}{WALS - Relation bw Order of Obj. \& Verb and Adp. \& NP} \\
\hline
2 - Vowel Quality \textit{(vowel)} & Phonology & Phonology & phonological rules based on vowel quality or class. & L\_Ulwa & \href{https://ecampusontario.pressbooks.pub/essentialsoflinguistics2/chapter/3-5-describing-vowels/}{Essentials of Linguistics: Describing vowels} & \href{https://wals.info/chapter/2}{WALS - Vowel Qual.} \\
\hline
10 - Vowel Nasalisation \textit{(nasal)} & Phonology & Phonology & phonological rules based on nasal consonants or sounds. & L\_Coptic & \href{https://www.oxfordbibliographies.com/display/document/obo-9780199772810/obo-9780199772810-0205.xml}{Nasals and Nasalization} & \href{https://wals.info/chapter/10}{WALS - Vowel Nasal.} \\
\hline

\end{tabular}%
}
\caption{Linguistic Feature tags used, their respective categorisations, definitions, example problem in the dataset, citation for the feature's presence in the example problem language, and related research on it.}
\label{tab:feat-def}
\end{table*}

\label{app:featdefs}

\section{Correlation Between Attributes and Model Scores}
See Tab. \ref{tab:feat_vs_corr_full}.

\section{Annotation Guidelines \& Inter-Annotator Agreement}
\label{app:ann}

Annotation guidelines are as follows:
\begin{itemize}
    \item Solve the problem yourself if the solution is not detailed enough to list out all the features used in the problem
    \item Identify the linguistic features that \textbf{are being tested} or borrow them from the solution (if present), map them to the finer features list (add if it is not already there).
    \item If a new feature was added in the last step, map it into WALS categorisation and broad categorisation (based on general definition).
    \item Mark 0 for the feature in the data column if there are data points that can be removed from the problem without affecting the solvability for that feature, 1 otherwise
    \item If the feature is something that English also has/uses, mark 1. If English has a similar feature to the same end, mark 1 there instead
    \item Use references linked in the last two columns for any confusion (here look at App. D)
\end{itemize}

See Tab. \ref{tab:ft_keep_agreement}. for inter-annotator agreement scores.

\begin{table}[]
\centering
\resizebox{\columnwidth}{!}{%
\begin{tabular}{@{}
>{\columncolor[HTML]{C0C0C0}}c lr
>{\columncolor[HTML]{C0C0C0}}c lr
>{\columncolor[HTML]{C0C0C0}}c lr
>{\columncolor[HTML]{C0C0C0}}c lr
>{\columncolor[HTML]{C0C0C0}}c lr@{}}
\toprule
\textbf{\#} & \multicolumn{1}{c}{\textbf{Feature}} & \multicolumn{1}{c}{\textbf{$\kappa$}} & \textbf{\#} & \multicolumn{1}{c}{\textbf{Feature}} & \multicolumn{1}{c}{\textbf{$\kappa$}} & \textbf{\#} & \multicolumn{1}{c}{\textbf{Feature}} & \multicolumn{1}{c}{\textbf{$\kappa$}} & \textbf{\#} & \multicolumn{1}{c}{\textbf{Feature}} & \multicolumn{1}{c}{\textbf{$\kappa$}} & \textbf{\#} & \multicolumn{1}{c}{\textbf{Feature}} & \multicolumn{1}{c}{\textbf{$\kappa$}} \\ \midrule
1 & \textit{adjnoun} & 1.00 & 11 & \textit{genfix} & 0.49 & 21 & \textit{nounadjc} & 1.00 & 31 & \textit{possfix} & 1.00 & 41 & \textit{svc} & 1.00 \\ \midrule
2 & \textit{alien} & 1.00 & 12 & \textit{intens} & 0.85 & 22 & \textit{nounposs} & 1.00 & 32 & \textit{posslinker} & 1.00 & 42 & \textit{svo} & 1.00 \\ \midrule
3 & \textit{anim} & 1.00 & 13 & \textit{miscfix} & 1.00 & 23 & \textit{nounprep} & 1.00 & 33 & \textit{prepc} & 0.66 & 43 & \textit{tensefix} & 0.84 \\ \midrule
4 & \textit{case} & 1.00 & 14 & \textit{miscphon} & 0.79 & 24 & \textit{num} & 1.00 & 34 & \textit{prepfix} & 1.00 & 44 & \textit{vso} & 1.00 \\ \midrule
5 & \textit{cls} & 0.90 & 15 & \textit{miscsyn} & 1.00 & 25 & \textit{numfix} & 0.94 & 35 & \textit{prepnoun} & 1.00 & 45 & \textit{tone} & 1.00 \\ \midrule
6 & \textit{comp} & 1.00 & 16 & \textit{modfix} & 1.00 & 26 & \textit{persfix} & 0.93 & 36 & \textit{quesfix} & 0.66 & 46 & \textit{transfix} & 0.91 \\ \midrule
7 & \textit{def} & 0.88 & 17 & \textit{moiety} & 1.00 & 27 & \textit{vowel} & 0.49 & 37 & \textit{redup} & 0.73 & 47 & \textit{voice} & 1.00 \\ \midrule
8 & \textit{nasal} & 0.66 & 18 & \textit{neg} & 1.00 & 28 & \textit{pl} & 1.00 & 38 & \textit{semfix} & 1.00 & 48 & \textit{vos} & 1.00 \\ \midrule
9 & \textit{dim} & 1.00 & 19 & \textit{negfix} & 1.00 & 29 & \textit{pron} & 0.88 & 39 & \textit{sov} & 0.95 & 49 & \textit{vs} & 1.00 \\ \midrule
10 & \textit{eli} & 1.00 & 20 & \textit{nounadj} & 1.00 & 30 & \textit{poss} & 1.00 & 40 & \textit{sv} & 1.00 & 50 & \textit{} & 0.00 \\ \bottomrule
\end{tabular}%
}
\caption{Pre-adjudication inter-annotator agreement (Cohen's $\kappa$) for marking feature presence in a puzzle. Note that after adjudication, all feature labelings have a perfect agreement ($\kappa$=1.0).}
\label{tab:ft_keep_agreement}
\end{table}

\section{Impact of Morpheme-Separated Input}
See Tab. \ref{tab:poc_qwise_gains_full}.

\begin{table*}[]
\centering
\resizebox{\textwidth}{!}{%
\begin{tabular}{@{}lrrrrrrrr@{}}
\toprule
\multirow{2}{*}{\textbf{Puzzle}} & \multicolumn{8}{c}{\textbf{Exact-Match Score (+Gain)}} \\ \cmidrule(l){2-9} 
 & \multicolumn{1}{c}{\texttt{Llama2-7B}} & \multicolumn{1}{c}{\texttt{Llama2-13B}} & \multicolumn{1}{c}{\texttt{Llama2-70B}} & \multicolumn{1}{c}{\texttt{Llama3-8B}} & \multicolumn{1}{c}{\texttt{Llama3-70B}} & \multicolumn{1}{c}{\texttt{Llama3.1-405B}} & \multicolumn{1}{c}{\texttt{GPT-4o}} & \multicolumn{1}{c}{\texttt{GPT-4}} \\ \midrule
\textit{L\_Ilokano} & 0.0 (+0.0) & 0.0 (+0.0) & 0.167 (+0.0) & 0.0 (-0.167) & 0.333 (+0.333) & 0.167 (+0.167) & 0.333 (+0.167) & 0.167 (+0.167) \\ \midrule
\textit{L\_Karelian} & 0.0 (-0.077) & 0.231 (-0.077) & 0.462 (+0.231) & 0.231 (-0.077) & 0.769 (+0.154) & 0.769 (-0.154) & 0.692 (-0.077) & 0.846 (+0.077) \\ \midrule
\textit{L\_Lardil} & 0.0 (+0.0) & 0.167 (-0.167) & 0.167 (+0.0) & 0.333 (+0.333) & 0.333 (+0.167) & 0.167 (+0.0) & 0.167 (-0.167) & 0.167 (+0.0) \\ \midrule
\textit{M\_Ayutla\_Mixe} & 0.0 (+0.0) & 0.0 (+0.0) & 0.0 (-0.25) & 0.0 (+0.0) & 0.0 (+0.0) & 0.0 (-0.25) & 0.0 (+0.0) & 0.0 (+0.0) \\ \midrule
\textit{M\_Bangime\_3} & 0.0 (+0.0) & 0.0 (+0.0) & 0.0 (+0.0) & 0.0 (+0.0) & 0.4 (+0.2) & 0.4 (+0.2) & 0.6 (+0.2) & 0.6 (+0.2) \\ \midrule
\textit{M\_Bangime\_5} & 0.0 (+0.0) & 0.4 (+0.4) & 0.4 (+0.2) & 0.2 (+0.2) & 0.6 (+0.2) & 0.8 (+0.4) & 0.6 (+0.2) & 0.8 (+0.2) \\ \midrule
\textit{M\_Guugu\_Yimithir} & 0.0 (-0.1) & 0.1 (+0.0) & 0.3 (+0.1) & 0.1 (-0.2) & 0.7 (+0.2) & 0.7 (+0.5) & 0.7 (+0.2) & 0.6 (+0.2) \\ \midrule
\textit{M\_Kutenai} & 0.2 (+0.0) & 0.2 (-0.4) & 0.4 (+0.2) & 0.4 (+0.0) & 1.0 (+0.6) & 0.8 (+0.0) & 0.6 (+0.2) & 0.6 (+0.2) \\ \midrule
\textit{M\_Totonac} & 0.167 (-0.167) & 0.167 (+0.167) & 0.167 (+0.167) & 0.0 (+0.0) & 0.333 (+0.167) & 0.333 (+0.167) & 0.333 (+0.167) & 0.167 (-0.333) \\ \bottomrule
\end{tabular}%
}
\caption{Puzzle-wise score gain on marking morpheme boundaries with whitespaces}
\label{tab:poc_qwise_gains_full}
\end{table*}

\begin{figure}
    \centering
    \vspace{-0.4cm}
    \includegraphics[width=0.6\linewidth]{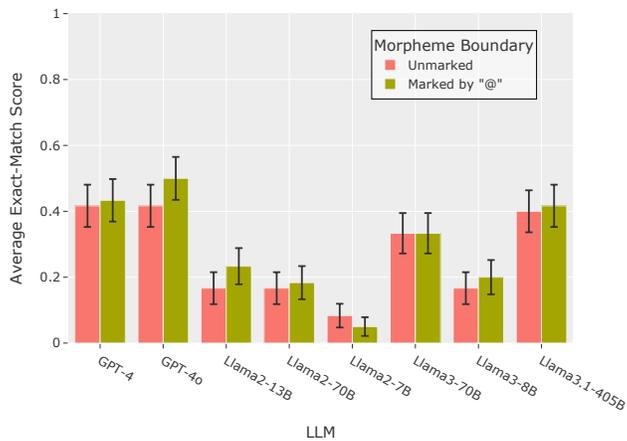}
    \vspace{-0.1cm}
    \caption{Performance change in models on making morpheme boundary explicit using \texttt{"@"}}
    \vspace{-0.1cm}
    \label{fig:poc-scores-for-AT}
\end{figure}

\begin{table*}[h]
\centering
\resizebox{\textwidth}{!}{%
\begin{tabular}{@{}llcccccccccc@{}}
\toprule
\multicolumn{2}{l}{\textit{}} & Llama2-7B & Llama2-13B & Llama2-70B & Mixtral-8x7B & Llama3-8B & Llama3-70B & Llama3.1-405B & GPT-3.5-Turbo & GPT-4o & GPT-4 \\ \midrule
\multicolumn{2}{l}{\textbf{Average Exact-Match Score}} & 0.074 & 0.143 & 0.224 & 0.271 & 0.269 & 0.381 & 0.405 & 0.248 & 0.488 & 0.408 \\ \midrule
\textbf{Attribute} & \textbf{Prompt Setting} & \multicolumn{10}{c}{} \\ \midrule
\multirow{7}{*}{\textit{avg\_eng}}
 & \textit{LingOlyStdPromptBuilder} & 0.031 & 0.053 & 0.388\textsuperscript{**} & 0.3\textsuperscript{*} & 0.491\textsuperscript{***} & 0.364\textsuperscript{**} & 0.276\textsuperscript{*} & 0.399\textsuperscript{**} & 0.258\textsuperscript{*} & 0.381\textsuperscript{**} \\
 & \textit{ModelingBasicCoTPromptBuilder} & -0.005 & 0.2 & 0.387\textsuperscript{**} & 0.401\textsuperscript{**} & 0.427\textsuperscript{***} & 0.385\textsuperscript{**} & 0.257\textsuperscript{*} & 0.52\textsuperscript{***} & 0.293\textsuperscript{*} & 0.379\textsuperscript{**} \\
 & \textit{ModelingFullCoTPromptBuilder} & 0.021 & 0.365\textsuperscript{**} & 0.246 & 0.339\textsuperscript{**} & 0.431\textsuperscript{***} & 0.343\textsuperscript{**} & 0.135 & 0.341\textsuperscript{**} & 0.26\textsuperscript{*} & 0.304\textsuperscript{*} \\
 & \textit{ModelingHandTunedPromptBuilder} & 0.026 & 0.46\textsuperscript{***} & 0.427\textsuperscript{***} & 0.483\textsuperscript{***} & 0.395\textsuperscript{**} & 0.408\textsuperscript{***} & 0.25\textsuperscript{*} & 0.369\textsuperscript{**} & 0.293\textsuperscript{*} & 0.387\textsuperscript{**} \\
 & \textit{ModelingMinimalPromptBuilder} & 0.16 & 0.385\textsuperscript{**} & 0.37\textsuperscript{**} & 0.385\textsuperscript{**} & 0.316\textsuperscript{*} & 0.468\textsuperscript{***} & 0.342\textsuperscript{**} & 0.406\textsuperscript{***} & 0.197 & 0.365\textsuperscript{**} \\
 & \textit{NullPromptBuilder} & -0.207 & -0.182 & 0.224 & 0.418\textsuperscript{***} & 0.354\textsuperscript{**} & 0.42\textsuperscript{***} & 0.117 & 0.259\textsuperscript{*} & 0.253\textsuperscript{*} & 0.296\textsuperscript{*} \\
 & \textit{Overall} & 0.031 & 0.053 & 0.388\textsuperscript{**} & 0.3\textsuperscript{*} & 0.491\textsuperscript{***} & 0.364\textsuperscript{**} & 0.276\textsuperscript{*} & 0.399\textsuperscript{**} & 0.258\textsuperscript{*} & 0.381\textsuperscript{**} \\
\midrule
\multirow{8}{*}{\textit{avg\_eng\_sim}}
 & \textit{LingOlyStdPromptBuilder} & -0.01 & -0.179 & -0.283\textsuperscript{*} & -0.206 & -0.385\textsuperscript{**} & -0.265\textsuperscript{*} & -0.146 & -0.216 & -0.092 & -0.224 \\
 & \textit{ModelingBasicCoTPromptBuilder} & -0.055 & -0.239 & -0.342\textsuperscript{**} & -0.331\textsuperscript{**} & -0.361\textsuperscript{**} & -0.301\textsuperscript{*} & -0.133 & -0.398\textsuperscript{**} & -0.148 & -0.283\textsuperscript{*} \\
 & \textit{ModelingFullCoTPromptBuilder} & -0.194 & -0.398\textsuperscript{**} & -0.22 & -0.24 & -0.371\textsuperscript{**} & -0.242 & 0.002 & -0.098 & -0.145 & -0.203 \\
 & \textit{ModelingHandTunedPromptBuilder} & -0.211 & -0.405\textsuperscript{***} & -0.307\textsuperscript{*} & -0.368\textsuperscript{**} & -0.25\textsuperscript{*} & -0.386\textsuperscript{**} & -0.084 & -0.179 & -0.15 & -0.304\textsuperscript{*} \\
 & \textit{ModelingMinimalPromptBuilder} & -0.17 & -0.272\textsuperscript{*} & -0.264\textsuperscript{*} & -0.271\textsuperscript{*} & -0.188 & -0.318\textsuperscript{*} & -0.218 & -0.301\textsuperscript{*} & -0.021 & -0.248\textsuperscript{*} \\
 & \textit{NullPromptBuilder} & -0.059 & -0.059 & -0.102 & -0.34\textsuperscript{**} & -0.348\textsuperscript{**} & -0.248\textsuperscript{*} & -0.006 & -0.179 & 0.035 & -0.099 \\
 & \textit{Overall} & -0.01 & -0.179 & -0.283\textsuperscript{*} & -0.206 & -0.385\textsuperscript{**} & -0.265\textsuperscript{*} & -0.146 & -0.216 & -0.092 & -0.224 \\ \midrule
\multirow{8}{*}{\textit{count\_exacts}}
 & \textit{LingOlyStdPromptBuilder} & -0.144 & -0.198 & -0.448\textsuperscript{***} & -0.466\textsuperscript{***} & -0.467\textsuperscript{***} & -0.55\textsuperscript{***} & -0.592\textsuperscript{***} & -0.543\textsuperscript{***} & -0.572\textsuperscript{***} & -0.547\textsuperscript{***} \\
 & \textit{ModelingBasicCoTPromptBuilder} & -0.238 & -0.287\textsuperscript{*} & -0.454\textsuperscript{***} & -0.477\textsuperscript{***} & -0.46\textsuperscript{***} & -0.539\textsuperscript{***} & -0.615\textsuperscript{***} & -0.495\textsuperscript{***} & -0.603\textsuperscript{***} & -0.533\textsuperscript{***} \\
 & \textit{ModelingFullCoTPromptBuilder} & -0.208 & -0.335\textsuperscript{**} & -0.457\textsuperscript{***} & -0.502\textsuperscript{***} & -0.451\textsuperscript{***} & -0.569\textsuperscript{***} & -0.674\textsuperscript{***} & -0.617\textsuperscript{***} & -0.577\textsuperscript{***} & -0.583\textsuperscript{***} \\
 & \textit{ModelingHandTunedPromptBuilder} & -0.17 & -0.38\textsuperscript{**} & -0.457\textsuperscript{***} & -0.533\textsuperscript{***} & -0.496\textsuperscript{***} & -0.512\textsuperscript{***} & -0.643\textsuperscript{***} & -0.623\textsuperscript{***} & -0.619\textsuperscript{***} & -0.511\textsuperscript{***} \\
 & \textit{ModelingMinimalPromptBuilder} & -0.271\textsuperscript{*} & -0.399\textsuperscript{**} & -0.472\textsuperscript{***} & -0.534\textsuperscript{***} & -0.537\textsuperscript{***} & -0.53\textsuperscript{***} & -0.359\textsuperscript{**} & -0.504\textsuperscript{***} & -0.649\textsuperscript{***} & -0.597\textsuperscript{***} \\
 & \textit{NullPromptBuilder} & -0.064 & -0.09 & -0.414\textsuperscript{***} & -0.481\textsuperscript{***} & -0.279\textsuperscript{*} & -0.544\textsuperscript{***} & -0.346\textsuperscript{**} & -0.376\textsuperscript{**} & -0.668\textsuperscript{***} & -0.612\textsuperscript{***} \\
 & \textit{Overall} & -0.144 & -0.198 & -0.448\textsuperscript{***} & -0.466\textsuperscript{***} & -0.467\textsuperscript{***} & -0.55\textsuperscript{***} & -0.592\textsuperscript{***} & -0.543\textsuperscript{***} & -0.572\textsuperscript{***} & -0.547\textsuperscript{***} \\ \midrule
\multirow{8}{*}{\textit{ft\_count}}
 & \textit{LingOlyStdPromptBuilder} & -0.17 & -0.263\textsuperscript{*} & -0.497\textsuperscript{***} & -0.544\textsuperscript{***} & -0.543\textsuperscript{***} & -0.609\textsuperscript{***} & -0.648\textsuperscript{***} & -0.548\textsuperscript{***} & -0.621\textsuperscript{***} & -0.567\textsuperscript{***} \\
 & \textit{ModelingBasicCoTPromptBuilder} & -0.303\textsuperscript{*} & -0.38\textsuperscript{**} & -0.531\textsuperscript{***} & -0.521\textsuperscript{***} & -0.523\textsuperscript{***} & -0.594\textsuperscript{***} & -0.67\textsuperscript{***} & -0.491\textsuperscript{***} & -0.627\textsuperscript{***} & -0.566\textsuperscript{***} \\
 & \textit{ModelingFullCoTPromptBuilder} & -0.25\textsuperscript{*} & -0.415\textsuperscript{***} & -0.509\textsuperscript{***} & -0.531\textsuperscript{***} & -0.541\textsuperscript{***} & -0.635\textsuperscript{***} & -0.742\textsuperscript{***} & -0.631\textsuperscript{***} & -0.636\textsuperscript{***} & -0.621\textsuperscript{***} \\
 & \textit{ModelingHandTunedPromptBuilder} & -0.254\textsuperscript{*} & -0.464\textsuperscript{***} & -0.523\textsuperscript{***} & -0.574\textsuperscript{***} & -0.566\textsuperscript{***} & -0.581\textsuperscript{***} & -0.69\textsuperscript{***} & -0.631\textsuperscript{***} & -0.659\textsuperscript{***} & -0.556\textsuperscript{***} \\
 & \textit{ModelingMinimalPromptBuilder} & -0.284\textsuperscript{*} & -0.448\textsuperscript{***} & -0.517\textsuperscript{***} & -0.585\textsuperscript{***} & -0.627\textsuperscript{***} & -0.57\textsuperscript{***} & -0.389\textsuperscript{**} & -0.545\textsuperscript{***} & -0.709\textsuperscript{***} & -0.631\textsuperscript{***} \\
 & \textit{NullPromptBuilder} & -0.124 & -0.148 & -0.479\textsuperscript{***} & -0.518\textsuperscript{***} & -0.382\textsuperscript{**} & -0.599\textsuperscript{***} & -0.403\textsuperscript{***} & -0.326\textsuperscript{**} & -0.7\textsuperscript{***} & -0.617\textsuperscript{***} \\
 & \textit{Overall} & -0.17 & -0.263\textsuperscript{*} & -0.497\textsuperscript{***} & -0.544\textsuperscript{***} & -0.543\textsuperscript{***} & -0.609\textsuperscript{***} & -0.648\textsuperscript{***} & -0.548\textsuperscript{***} & -0.621\textsuperscript{***} & -0.567\textsuperscript{***} \\ \midrule
\multirow{8}{*}{\textit{s\_Morphology}}
 & \textit{LingOlyStdPromptBuilder} & -0.156 & -0.261\textsuperscript{*} & -0.482\textsuperscript{***} & -0.558\textsuperscript{***} & -0.554\textsuperscript{***} & -0.604\textsuperscript{***} & -0.652\textsuperscript{***} & -0.514\textsuperscript{***} & -0.595\textsuperscript{***} & -0.507\textsuperscript{***} \\
 & \textit{ModelingBasicCoTPromptBuilder} & -0.281\textsuperscript{*} & -0.377\textsuperscript{**} & -0.507\textsuperscript{***} & -0.52\textsuperscript{***} & -0.494\textsuperscript{***} & -0.562\textsuperscript{***} & -0.666\textsuperscript{***} & -0.444\textsuperscript{***} & -0.57\textsuperscript{***} & -0.523\textsuperscript{***} \\
 & \textit{ModelingFullCoTPromptBuilder} & -0.234 & -0.421\textsuperscript{***} & -0.507\textsuperscript{***} & -0.535\textsuperscript{***} & -0.544\textsuperscript{***} & -0.609\textsuperscript{***} & -0.729\textsuperscript{***} & -0.618\textsuperscript{***} & -0.618\textsuperscript{***} & -0.585\textsuperscript{***} \\
 & \textit{ModelingHandTunedPromptBuilder} & -0.296\textsuperscript{*} & -0.469\textsuperscript{***} & -0.497\textsuperscript{***} & -0.569\textsuperscript{***} & -0.565\textsuperscript{***} & -0.554\textsuperscript{***} & -0.688\textsuperscript{***} & -0.581\textsuperscript{***} & -0.643\textsuperscript{***} & -0.523\textsuperscript{***} \\
 & \textit{ModelingMinimalPromptBuilder} & -0.262\textsuperscript{*} & -0.426\textsuperscript{***} & -0.528\textsuperscript{***} & -0.585\textsuperscript{***} & -0.614\textsuperscript{***} & -0.565\textsuperscript{***} & -0.405\textsuperscript{***} & -0.564\textsuperscript{***} & -0.707\textsuperscript{***} & -0.605\textsuperscript{***} \\
 & \textit{NullPromptBuilder} & -0.094 & -0.106 & -0.477\textsuperscript{***} & -0.543\textsuperscript{***} & -0.377\textsuperscript{**} & -0.611\textsuperscript{***} & -0.399\textsuperscript{**} & -0.283\textsuperscript{*} & -0.691\textsuperscript{***} & -0.581\textsuperscript{***} \\
 & \textit{Overall} & -0.219\textsuperscript{***} & -0.348\textsuperscript{***} & -0.496\textsuperscript{***} & -0.546\textsuperscript{***} & -0.522\textsuperscript{***} & -0.583\textsuperscript{***} & -0.572\textsuperscript{***} & -0.483\textsuperscript{***} & -0.637\textsuperscript{***} & -0.553\textsuperscript{***} \\ \midrule
\multirow{8}{*}{\textit{s\_Phonology}}
 & \textit{LingOlyStdPromptBuilder} & 0.001 & -0.05 & -0.16 & -0.25\textsuperscript{*} & -0.218 & -0.227 & -0.243 & -0.235 & -0.341\textsuperscript{**} & -0.273\textsuperscript{*} \\
 & \textit{ModelingBasicCoTPromptBuilder} & -0.044 & -0.017 & -0.225 & -0.233 & -0.216 & -0.221 & -0.228 & -0.2 & -0.289\textsuperscript{*} & -0.227 \\
 & \textit{ModelingFullCoTPromptBuilder} & -0.039 & -0.093 & -0.226 & -0.191 & -0.169 & -0.253\textsuperscript{*} & -0.215 & -0.301\textsuperscript{*} & -0.275\textsuperscript{*} & -0.326\textsuperscript{**} \\
 & \textit{ModelingHandTunedPromptBuilder} & -0.035 & -0.137 & -0.168 & -0.218 & -0.288\textsuperscript{*} & -0.179 & -0.257\textsuperscript{*} & -0.272\textsuperscript{*} & -0.3\textsuperscript{*} & -0.259\textsuperscript{*} \\
 & \textit{ModelingMinimalPromptBuilder} & -0.172 & -0.175 & -0.203 & -0.18 & -0.307\textsuperscript{*} & -0.215 & -0.207 & -0.2 & -0.341\textsuperscript{**} & -0.281\textsuperscript{*} \\
 & \textit{NullPromptBuilder} & -0.063 & -0.012 & -0.178 & -0.15 & -0.156 & -0.251\textsuperscript{*} & -0.255\textsuperscript{*} & -0.068 & -0.355\textsuperscript{**} & -0.264\textsuperscript{*} \\
 & \textit{Overall} & 0.001 & -0.05 & -0.16 & -0.25\textsuperscript{*} & -0.218 & -0.227 & -0.243 & -0.235 & -0.341\textsuperscript{**} & -0.273\textsuperscript{*} \\ \midrule
\multirow{8}{*}{\textit{s\_Semantics}}
 & \textit{LingOlyStdPromptBuilder} & -0.084 & 0.028 & -0.31\textsuperscript{*} & -0.215 & -0.25\textsuperscript{*} & -0.329\textsuperscript{**} & -0.291\textsuperscript{*} & -0.354\textsuperscript{**} & -0.331\textsuperscript{**} & -0.347\textsuperscript{**} \\
 & \textit{ModelingBasicCoTPromptBuilder} & -0.02 & -0.136 & -0.199 & -0.261\textsuperscript{*} & -0.258\textsuperscript{*} & -0.289\textsuperscript{*} & -0.296\textsuperscript{*} & -0.333\textsuperscript{**} & -0.349\textsuperscript{**} & -0.297\textsuperscript{*} \\
 & \textit{ModelingFullCoTPromptBuilder} & 0.047 & -0.101 & -0.206 & -0.314\textsuperscript{*} & -0.229 & -0.302\textsuperscript{*} & -0.311\textsuperscript{*} & -0.369\textsuperscript{**} & -0.304\textsuperscript{*} & -0.252\textsuperscript{*} \\
 & \textit{ModelingHandTunedPromptBuilder} & 0.071 & -0.181 & -0.282\textsuperscript{*} & -0.32\textsuperscript{**} & -0.224 & -0.239 & -0.322\textsuperscript{**} & -0.367\textsuperscript{**} & -0.342\textsuperscript{**} & -0.277\textsuperscript{*} \\
 & \textit{ModelingMinimalPromptBuilder} & -0.084 & -0.202 & -0.271\textsuperscript{*} & -0.293\textsuperscript{*} & -0.212 & -0.302\textsuperscript{*} & -0.115 & -0.258\textsuperscript{*} & -0.341\textsuperscript{**} & -0.336\textsuperscript{**} \\
 & \textit{NullPromptBuilder} & 0.142 & 0.078 & -0.227 & -0.298\textsuperscript{*} & -0.157 & -0.305\textsuperscript{*} & -0.067 & -0.249\textsuperscript{*} & -0.406\textsuperscript{***} & -0.434\textsuperscript{***} \\
 & \textit{Overall} & -0.084 & 0.028 & -0.31\textsuperscript{*} & -0.215 & -0.25\textsuperscript{*} & -0.329\textsuperscript{**} & -0.291\textsuperscript{*} & -0.354\textsuperscript{**} & -0.331\textsuperscript{**} & -0.347\textsuperscript{**} \\ \midrule
\multirow{7}{*}{\textit{s\_Syntax}}
 & \textit{LingOlyStdPromptBuilder} & -0.083 & -0.183 & -0.073 & -0.079 & -0.074 & -0.089 & -0.118 & -0.083 & -0.087 & -0.134 \\
 & \textit{ModelingBasicCoTPromptBuilder} & -0.215 & -0.157 & -0.169 & -0.067 & -0.134 & -0.164 & -0.151 & -0.1 & -0.161 & -0.157 \\
 & \textit{ModelingFullCoTPromptBuilder} & -0.222 & -0.159 & -0.105 & -0.044 & -0.122 & -0.154 & -0.211 & -0.057 & -0.13 & -0.174 \\
 & \textit{ModelingHandTunedPromptBuilder} & -0.129 & -0.118 & -0.129 & -0.075 & -0.103 & -0.197 & -0.129 & -0.144 & -0.104 & -0.132 \\
 & \textit{ModelingMinimalPromptBuilder} & -0.101 & -0.125 & -0.047 & -0.107 & -0.169 & -0.086 & -0.061 & -0.061 & -0.095 & -0.117 \\
 & \textit{NullPromptBuilder} & -0.222 & -0.236 & -0.09 & -0.023 & -0.088 & -0.056 & -0.128 & -0.096 & -0.046 & -0.062 \\
 & \textit{Overall} & -0.083 & -0.183 & -0.073 & -0.079 & -0.074 & -0.089 & -0.118 & -0.083 & -0.087 & -0.134 \\ \midrule
\multirow{7}{*}{\textit{w\_Complex\_Sentences}}
 & \textit{LingOlyStdPromptBuilder} & -0.081 & -0.101 & -0.14 & -0.162 & -0.154 & -0.198 & -0.198 & -0.166 & -0.215 & -0.18 \\
 & \textit{ModelingBasicCoTPromptBuilder} & -0.098 & -0.116 & -0.151 & -0.157 & -0.156 & -0.199 & -0.214 & -0.143 & -0.181 & -0.192 \\
 & \textit{ModelingFullCoTPromptBuilder} & -0.098 & -0.13 & -0.164 & -0.164 & -0.159 & -0.192 & -0.23 & -0.188 & -0.177 & -0.205 \\
 & \textit{ModelingHandTunedPromptBuilder} & -0.108 & -0.138 & -0.16 & -0.165 & -0.169 & -0.191 & -0.202 & -0.183 & -0.16 & -0.185 \\
 & \textit{ModelingMinimalPromptBuilder} & -0.105 & -0.125 & -0.153 & -0.183 & -0.178 & -0.153 & -0.155 & -0.157 & -0.175 & -0.196 \\
 & \textit{NullPromptBuilder} & -0.082 & -0.088 & -0.149 & -0.143 & -0.13 & -0.17 & -0.13 & -0.085 & -0.229 & -0.204 \\
 & \textit{Overall} & -0.081 & -0.101 & -0.14 & -0.162 & -0.154 & -0.198 & -0.198 & -0.166 & -0.215 & -0.18 \\ \bottomrule
\end{tabular}%
}
\end{table*}

% I'M BREAKING HERE

\begin{table*}[h]
\centering
\resizebox{\textwidth}{!}{%
\begin{tabular}{@{}llcccccccccc@{}}
\toprule
\multirow{7}{*}{\textit{w\_Lexicon}}
 & \textit{LingOlyStdPromptBuilder} & -0.1 & -0.125 & -0.172 & -0.182 & -0.172 & -0.203 & -0.147 & -0.153 & -0.168 & -0.168 \\
 & \textit{ModelingBasicCoTPromptBuilder} & -0.121 & -0.118 & -0.187 & -0.194 & -0.174 & -0.164 & -0.157 & -0.158 & -0.191 & -0.184 \\
 & \textit{ModelingFullCoTPromptBuilder} & -0.085 & -0.139 & -0.182 & -0.18 & -0.197 & -0.182 & -0.182 & -0.184 & -0.154 & -0.146 \\
 & \textit{ModelingHandTunedPromptBuilder} & -0.093 & -0.149 & -0.177 & -0.187 & -0.179 & -0.151 & -0.158 & -0.209 & -0.171 & -0.19 \\
 & \textit{ModelingMinimalPromptBuilder} & -0.082 & -0.127 & -0.189 & -0.18 & -0.141 & -0.144 & -0.135 & -0.143 & -0.24 & -0.163 \\
 & \textit{NullPromptBuilder} & -0.064 & -0.076 & -0.184 & -0.158 & -0.137 & -0.128 & -0.024 & -0.053 & -0.153 & -0.237 \\
 & \textit{Overall} & -0.1 & -0.125 & -0.172 & -0.182 & -0.172 & -0.203 & -0.147 & -0.153 & -0.168 & -0.168 \\ \midrule
\multirow{8}{*}{\textit{w\_Morphology}}
 & \textit{LingOlyStdPromptBuilder} & -0.035 & -0.133 & -0.245 & -0.308\textsuperscript{*} & -0.293\textsuperscript{*} & -0.409\textsuperscript{***} & -0.395\textsuperscript{**} & -0.312\textsuperscript{*} & -0.401\textsuperscript{**} & -0.309\textsuperscript{*} \\
 & \textit{ModelingBasicCoTPromptBuilder} & -0.024 & -0.211 & -0.323\textsuperscript{**} & -0.277\textsuperscript{*} & -0.285\textsuperscript{*} & -0.4\textsuperscript{**} & -0.434\textsuperscript{***} & -0.348\textsuperscript{**} & -0.428\textsuperscript{***} & -0.339\textsuperscript{**} \\
 & \textit{ModelingFullCoTPromptBuilder} & -0.109 & -0.219 & -0.324\textsuperscript{**} & -0.302\textsuperscript{*} & -0.286\textsuperscript{*} & -0.396\textsuperscript{**} & -0.395\textsuperscript{**} & -0.415\textsuperscript{***} & -0.437\textsuperscript{***} & -0.383\textsuperscript{**} \\
 & \textit{ModelingHandTunedPromptBuilder} & -0.136 & -0.253\textsuperscript{*} & -0.302\textsuperscript{*} & -0.35\textsuperscript{**} & -0.306\textsuperscript{*} & -0.379\textsuperscript{**} & -0.396\textsuperscript{**} & -0.418\textsuperscript{***} & -0.408\textsuperscript{***} & -0.351\textsuperscript{**} \\
 & \textit{ModelingMinimalPromptBuilder} & -0.171 & -0.232 & -0.272\textsuperscript{*} & -0.364\textsuperscript{**} & -0.369\textsuperscript{**} & -0.365\textsuperscript{**} & -0.17 & -0.356\textsuperscript{**} & -0.424\textsuperscript{***} & -0.403\textsuperscript{***} \\
 & \textit{NullPromptBuilder} & 0.038 & 0.018 & -0.299\textsuperscript{*} & -0.343\textsuperscript{**} & -0.246\textsuperscript{*} & -0.359\textsuperscript{**} & -0.285\textsuperscript{*} & -0.235 & -0.401\textsuperscript{**} & -0.419\textsuperscript{***} \\
 & \textit{Overall} & -0.035 & -0.133 & -0.245 & -0.308\textsuperscript{*} & -0.293\textsuperscript{*} & -0.409\textsuperscript{***} & -0.395\textsuperscript{**} & -0.312\textsuperscript{*} & -0.401\textsuperscript{**} & -0.309\textsuperscript{*} \\ \midrule
\multirow{8}{*}{\textit{w\_Nominal\_Categories}}
 & \textit{LingOlyStdPromptBuilder} & -0.13 & -0.033 & -0.399\textsuperscript{**} & -0.377\textsuperscript{**} & -0.374\textsuperscript{**} & -0.397\textsuperscript{**} & -0.426\textsuperscript{***} & -0.422\textsuperscript{***} & -0.464\textsuperscript{***} & -0.413\textsuperscript{***} \\
 & \textit{ModelingBasicCoTPromptBuilder} & -0.16 & -0.202 & -0.282\textsuperscript{*} & -0.382\textsuperscript{**} & -0.329\textsuperscript{**} & -0.343\textsuperscript{**} & -0.433\textsuperscript{***} & -0.333\textsuperscript{**} & -0.45\textsuperscript{***} & -0.352\textsuperscript{**} \\
 & \textit{ModelingFullCoTPromptBuilder} & -0.015 & -0.206 & -0.285\textsuperscript{*} & -0.423\textsuperscript{***} & -0.334\textsuperscript{**} & -0.382\textsuperscript{**} & -0.472\textsuperscript{***} & -0.469\textsuperscript{***} & -0.378\textsuperscript{**} & -0.38\textsuperscript{**} \\
 & \textit{ModelingHandTunedPromptBuilder} & -0.038 & -0.278\textsuperscript{*} & -0.337\textsuperscript{**} & -0.423\textsuperscript{***} & -0.364\textsuperscript{**} & -0.31\textsuperscript{*} & -0.472\textsuperscript{***} & -0.417\textsuperscript{***} & -0.479\textsuperscript{***} & -0.357\textsuperscript{**} \\
 & \textit{ModelingMinimalPromptBuilder} & -0.053 & -0.282\textsuperscript{*} & -0.352\textsuperscript{**} & -0.414\textsuperscript{***} & -0.352\textsuperscript{**} & -0.399\textsuperscript{**} & -0.247\textsuperscript{*} & -0.377\textsuperscript{**} & -0.522\textsuperscript{***} & -0.448\textsuperscript{***} \\
 & \textit{NullPromptBuilder} & 0.062 & 0.027 & -0.272\textsuperscript{*} & -0.405\textsuperscript{***} & -0.215 & -0.416\textsuperscript{***} & -0.138 & -0.239 & -0.566\textsuperscript{***} & -0.478\textsuperscript{***} \\
 & \textit{Overall} & -0.13 & -0.033 & -0.399\textsuperscript{**} & -0.377\textsuperscript{**} & -0.374\textsuperscript{**} & -0.397\textsuperscript{**} & -0.426\textsuperscript{***} & -0.422\textsuperscript{***} & -0.464\textsuperscript{***} & -0.413\textsuperscript{***} \\ \midrule
\multirow{7}{*}{\textit{w\_Nominal\_Syntax}}
 & \textit{LingOlyStdPromptBuilder} & 0.225 & 0.034 & -0.033 & -0.057 & -0.087 & -0.111 & -0.131 & -0.182 & -0.122 & -0.152 \\
 & \textit{ModelingBasicCoTPromptBuilder} & 0.033 & 0.019 & -0.102 & 0.034 & -0.094 & -0.144 & -0.201 & -0.18 & -0.228 & -0.164 \\
 & \textit{ModelingFullCoTPromptBuilder} & 0 & 0.002 & -0.033 & -0.024 & -0.086 & -0.172 & -0.265\textsuperscript{*} & -0.221 & -0.184 & -0.197 \\
 & \textit{ModelingHandTunedPromptBuilder} & 0.175 & -0.013 & -0.116 & -0.069 & -0.065 & -0.153 & -0.213 & -0.276\textsuperscript{*} & -0.151 & -0.119 \\
 & \textit{ModelingMinimalPromptBuilder} & -0.056 & -0.055 & -0.019 & -0.109 & -0.17 & -0.129 & 0.004 & -0.146 & -0.154 & -0.146 \\
 & \textit{NullPromptBuilder} & 0.149 & 0.115 & -0.142 & -0.029 & 0.089 & -0.167 & -0.225 & -0.157 & -0.187 & -0.171 \\
 & \textit{Overall} & 0.225 & 0.034 & -0.033 & -0.057 & -0.087 & -0.111 & -0.131 & -0.182 & -0.122 & -0.152 \\ \midrule
\multirow{7}{*}{\textit{w\_Phonology}}
 & \textit{LingOlyStdPromptBuilder} & 0.021 & -0.027 & -0.132 & -0.221 & -0.19 & -0.188 & -0.204 & -0.243 & -0.327\textsuperscript{**} & -0.24 \\
 & \textit{ModelingBasicCoTPromptBuilder} & -0.022 & 0.011 & -0.197 & -0.203 & -0.186 & -0.181 & -0.24 & -0.173 & -0.273\textsuperscript{*} & -0.218 \\
 & \textit{ModelingFullCoTPromptBuilder} & -0.016 & -0.065 & -0.195 & -0.158 & -0.137 & -0.216 & -0.226 & -0.268\textsuperscript{*} & -0.26\textsuperscript{*} & -0.289\textsuperscript{*} \\
 & \textit{ModelingHandTunedPromptBuilder} & -0.009 & -0.108 & -0.135 & -0.186 & -0.259\textsuperscript{*} & -0.139 & -0.241 & -0.239 & -0.251\textsuperscript{*} & -0.224 \\
 & \textit{ModelingMinimalPromptBuilder} & -0.153 & -0.152 & -0.173 & -0.177 & -0.276\textsuperscript{*} & -0.178 & -0.269\textsuperscript{*} & -0.17 & -0.322\textsuperscript{**} & -0.244 \\
 & \textit{NullPromptBuilder} & -0.046 & 0.009 & -0.149 & -0.121 & -0.13 & -0.22 & -0.233 & -0.05 & -0.344\textsuperscript{**} & -0.255\textsuperscript{*} \\
 & \textit{Overall} & 0.021 & -0.027 & -0.132 & -0.221 & -0.19 & -0.188 & -0.204 & -0.243 & -0.327\textsuperscript{**} & -0.24 \\ \midrule
\multirow{7}{*}{\textit{w\_Simple\_Clauses}}
 & \textit{LingOlyStdPromptBuilder} & -0.207 & -0.33\textsuperscript{**} & -0.467\textsuperscript{***} & -0.518\textsuperscript{***} & -0.553\textsuperscript{***} & -0.566\textsuperscript{***} & -0.625\textsuperscript{***} & -0.443\textsuperscript{***} & -0.47\textsuperscript{***} & -0.436\textsuperscript{***} \\
 & \textit{ModelingBasicCoTPromptBuilder} & -0.316\textsuperscript{*} & -0.417\textsuperscript{***} & -0.5\textsuperscript{***} & -0.496\textsuperscript{***} & -0.486\textsuperscript{***} & -0.551\textsuperscript{***} & -0.57\textsuperscript{***} & -0.389\textsuperscript{**} & -0.412\textsuperscript{***} & -0.478\textsuperscript{***} \\
 & \textit{ModelingFullCoTPromptBuilder} & -0.277\textsuperscript{*} & -0.457\textsuperscript{***} & -0.505\textsuperscript{***} & -0.485\textsuperscript{***} & -0.552\textsuperscript{***} & -0.583\textsuperscript{***} & -0.659\textsuperscript{***} & -0.529\textsuperscript{***} & -0.572\textsuperscript{***} & -0.508\textsuperscript{***} \\
 & \textit{ModelingHandTunedPromptBuilder} & -0.397\textsuperscript{**} & -0.509\textsuperscript{***} & -0.502\textsuperscript{***} & -0.517\textsuperscript{***} & -0.56\textsuperscript{***} & -0.574\textsuperscript{***} & -0.62\textsuperscript{***} & -0.496\textsuperscript{***} & -0.566\textsuperscript{***} & -0.469\textsuperscript{***} \\
 & \textit{ModelingMinimalPromptBuilder} & -0.362\textsuperscript{**} & -0.447\textsuperscript{***} & -0.559\textsuperscript{***} & -0.529\textsuperscript{***} & -0.618\textsuperscript{***} & -0.528\textsuperscript{***} & -0.356\textsuperscript{**} & -0.523\textsuperscript{***} & -0.583\textsuperscript{***} & -0.517\textsuperscript{***} \\
 & \textit{NullPromptBuilder} & -0.195 & -0.195 & -0.486\textsuperscript{***} & -0.486\textsuperscript{***} & -0.408\textsuperscript{***} & -0.578\textsuperscript{***} & -0.424\textsuperscript{***} & -0.244 & -0.559\textsuperscript{***} & -0.447\textsuperscript{***} \\
 & \textit{Overall} & -0.207 & -0.33\textsuperscript{**} & -0.467\textsuperscript{***} & -0.518\textsuperscript{***} & -0.553\textsuperscript{***} & -0.566\textsuperscript{***} & -0.625\textsuperscript{***} & -0.443\textsuperscript{***} & -0.47\textsuperscript{***} & -0.436\textsuperscript{***} \\ \midrule
\multirow{8}{*}{\textit{w\_Verbal Categories}}
 & \textit{LingOlyStdPromptBuilder} & -0.194 & -0.243 & -0.282\textsuperscript{*} & -0.333\textsuperscript{**} & -0.333\textsuperscript{**} & -0.39\textsuperscript{**} & -0.422\textsuperscript{***} & -0.308\textsuperscript{*} & -0.392\textsuperscript{**} & -0.354\textsuperscript{**} \\
 & \textit{ModelingBasicCoTPromptBuilder} & -0.234 & -0.278\textsuperscript{*} & -0.343\textsuperscript{**} & -0.346\textsuperscript{**} & -0.322\textsuperscript{**} & -0.372\textsuperscript{**} & -0.421\textsuperscript{***} & -0.247\textsuperscript{*} & -0.356\textsuperscript{**} & -0.333\textsuperscript{**} \\
 & \textit{ModelingFullCoTPromptBuilder} & -0.235 & -0.295\textsuperscript{*} & -0.35\textsuperscript{**} & -0.328\textsuperscript{**} & -0.35\textsuperscript{**} & -0.38\textsuperscript{**} & -0.508\textsuperscript{***} & -0.357\textsuperscript{**} & -0.404\textsuperscript{***} & -0.359\textsuperscript{**} \\
 & \textit{ModelingHandTunedPromptBuilder} & -0.26\textsuperscript{*} & -0.3\textsuperscript{*} & -0.326\textsuperscript{**} & -0.34\textsuperscript{**} & -0.355\textsuperscript{**} & -0.362\textsuperscript{**} & -0.417\textsuperscript{***} & -0.347\textsuperscript{**} & -0.396\textsuperscript{**} & -0.299\textsuperscript{*} \\
 & \textit{ModelingMinimalPromptBuilder} & -0.252\textsuperscript{*} & -0.279\textsuperscript{*} & -0.355\textsuperscript{**} & -0.341\textsuperscript{**} & -0.375\textsuperscript{**} & -0.345\textsuperscript{**} & -0.328\textsuperscript{**} & -0.292\textsuperscript{*} & -0.43\textsuperscript{***} & -0.373\textsuperscript{**} \\
 & \textit{NullPromptBuilder} & -0.15 & -0.15 & -0.288\textsuperscript{*} & -0.343\textsuperscript{**} & -0.283\textsuperscript{*} & -0.36\textsuperscript{**} & -0.244 & -0.203 & -0.443\textsuperscript{***} & -0.353\textsuperscript{**} \\
 & \textit{Overall} & -0.194 & -0.243 & -0.282\textsuperscript{*} & -0.333\textsuperscript{**} & -0.333\textsuperscript{**} & -0.39\textsuperscript{**} & -0.422\textsuperscript{***} & -0.308\textsuperscript{*} & -0.392\textsuperscript{**} & -0.354\textsuperscript{**} \\ \midrule
\multirow{7}{*}{\textit{w\_Word\_Order}}
 & \textit{LingOlyStdPromptBuilder} & -0.254\textsuperscript{*} & -0.271\textsuperscript{*} & -0.034 & -0.019 & 0.02 & 0.003 & 0 & 0.101 & 0.052 & -0.005 \\
 & \textit{ModelingBasicCoTPromptBuilder} & -0.313\textsuperscript{*} & -0.221 & -0.12 & -0.082 & -0.081 & -0.058 & 0.002 & 0.032 & 0.021 & -0.045 \\ \midrule
& \textit{ModelingFullCoTPromptBuilder} & -0.333\textsuperscript{**} & -0.207 & -0.092 & -0.011 & -0.068 & -0.046 & 0.002 & 0.172 & 0.009 & -0.049 \\
 & \textit{ModelingHandTunedPromptBuilder} & -0.319\textsuperscript{*} & -0.116 & -0.027 & -0.016 & -0.037 & -0.086 & 0.034 & 0.079 & 0.028 & -0.067 \\
 & \textit{ModelingMinimalPromptBuilder} & -0.054 & -0.085 & 0.001 & -0.005 & -0.04 & 0.034 & -0.034 & 0.06 & 0.041 & -0.003 \\
 & \textit{NullPromptBuilder} & -0.433\textsuperscript{***} & -0.432\textsuperscript{***} & 0.02 & 0.031 & -0.159 & 0.087 & 0.037 & 0.047 & 0.166 & 0.095 \\
 & \textit{Overall} & -0.254\textsuperscript{*} & -0.271\textsuperscript{*} & -0.034 & -0.019 & 0.02 & 0.003 & 0 & 0.101 & 0.052 & -0.005 \\ \bottomrule
\end{tabular}
}

\caption{Pearson correlation values between exact-match scores and feature values for all features against all model and prompt-setting combinations. Correlation values marked with ``*", ``**", and ``***" have a corresponding $p$-value less than 0.05, 0.01, and 0.001, respectively.}
\label{tab:feat_vs_corr_full}
\end{table*}

\end{document}